%% file: icml2024_main_CR.tex
\documentclass{article}

\usepackage{microtype}
\usepackage{graphicx}
\usepackage{subfigure}
\usepackage{booktabs} % for professional tables

\usepackage{hyperref}

\usepackage[accepted]{icml2024_ArXiv}

\usepackage{amsmath}
\usepackage{amssymb}
\usepackage{mathtools}
\usepackage{amsthm}

\usepackage[capitalize,noabbrev]{cleveref}

\theoremstyle{plain}
\usepackage[textsize=tiny]{todonotes}

\usepackage{xcolor}         % colors
\usepackage{graphicx}
\usepackage{epsf}
\usepackage{fancyhdr}
\usepackage{graphics}
\usepackage{psfrag}

\usepackage{wrapfig}
\usepackage{float}
\usepackage{algorithm}
\usepackage{algorithmic}

\usepackage{color}

\usepackage{bm,bbm}
\usepackage{multirow}
\usepackage{balance}

\def\calGS{\mathcal{GS}}

\def\OrliczSobolevPhi{W\!L_{\Phi}}
\def\OrliczSobolevPsi{W\!L_{\Psi}}

\def\OrliczSobolevPsiOne{W\!L_{\Psi_1}}
\def\OrliczSobolevPsiTwo{W\!L_{\Psi_2}}

\usepackage{verbatim}

\input{comments}

\icmltitlerunning{Generalized Sobolev Transport for Probability Measures on a Graph}

\begin{document}

\twocolumn[
\icmltitle{Generalized Sobolev Transport for Probability Measures on a Graph}



\icmlsetsymbol{equal}{*}

\begin{icmlauthorlist}
\icmlauthor{Tam Le}{equal,ism,riken}
\icmlauthor{Truyen Nguyen}{equal,akron}
\icmlauthor{Kenji Fukumizu}{ism}

%%\icmlauthor{Firstname4 Lastname4}{sch}
%%\icmlauthor{Firstname5 Lastname5}{yyy}
%%\icmlauthor{Firstname6 Lastname6}{sch,yyy,comp}
%%\icmlauthor{Firstname7 Lastname7}{comp}
%%%\icmlauthor{}{sch}
%%\icmlauthor{Firstname8 Lastname8}{sch}
%%\icmlauthor{Firstname8 Lastname8}{yyy,comp}
\end{icmlauthorlist}

%%\icmlaffiliation{yyy}{Department of XXX, University of YYY, Location, Country}
%%\icmlaffiliation{comp}{Company Name, Location, Country}
%%\icmlaffiliation{sch}{School of ZZZ, Institute of WWW, Location, Country}

\icmlaffiliation{ism}{Department of Advanced Data Science, The Institute of Statistical Mathematics (ISM), Tokyo, Japan}
\icmlaffiliation{riken}{RIKEN AIP, Tokyo, Japan}
\icmlaffiliation{akron}{The University of Akron, Ohio, US}

\icmlcorrespondingauthor{Tam Le}{tam@ism.ac.jp}
%\icmlcorrespondingauthor{Firstname2 Lastname2}{first2.last2@www.uk}

\icmlkeywords{Scalability, Generalized Sobolev Transport, Measures on a Graph, Lp geometric structure, Orlicz geometric structure, Sobolev Transport, Orlicz-Wasserstein}

\vskip 0.3in
]



\printAffiliationsAndNotice{\icmlEqualContribution} % otherwise use the standard text.

\begin{abstract}
We study the optimal transport (OT) problem for measures supported on a graph metric space. Recently,~\citet{le2022st} leverage the graph structure and propose a variant of OT, namely Sobolev transport (ST), which yields a closed-form expression for a fast computation. However, ST is essentially coupled with the $L^p$ geometric structure within its definition which  makes it  nontrivial to utilize ST for other prior structures. In contrast, the classic OT has the flexibility to adapt to various geometric structures by modifying the underlying cost function. An important instance is the Orlicz-Wasserstein (OW) which moves beyond the $L^p$ structure by leveraging the \emph{Orlicz geometric structure}. Comparing to the usage of standard $p$-order Wasserstein, OW remarkably helps to advance certain machine learning approaches. Nevertheless, OW brings up a new challenge on its computation due to its two-level optimization formulation. In this work, we leverage a specific class of convex functions for Orlicz structure  to propose the generalized Sobolev transport (GST). GST encompasses the ST as its special case, and  can be utilized for prior structures  beyond  the $L^p$ geometry. In connection with the OW, we show that one only needs to simply solve a univariate optimization problem to compute the GST, unlike the complex two-level optimization problem in OW. We empirically illustrate that GST is several-order faster than the OW. Moreover, we provide preliminary evidences on the advantages of GST for document classification and for several tasks in topological data analysis.
\end{abstract}

%%%%%%%%%%%%%%%%%%%%%%%%%%%%%%%%%%%%%%%%%%%%
%%%%%%%%%%%%%%%%%%%%%%%%%%%%%%%%%%%%%%%%%%%%
\section{Introduction}
\label{sec:intro}

%le2021robust
%nguyen2024quasi

Optimal transport (OT) is a natural geometry to compare probability measures~\citep{villani2008optimal, peyre2019computational}. Intuitively, OT lifts the cost metric on support data points of input measures to the distance for the measures by minimizing the cost to transport a measure into the other. OT has been applied on many applications in various research fields, e.g., in machine learning~\citep{nadjahi2019asymptotic, titouan2019optimal, janati2020entropic, mukherjee2021outlier, altschuler2021averaging, fatras2021unbalanced,  scetbon2021low, le2021ept, le2021flow, le2021adversarial, liu2021lsmi,  nguyen2021robust, nguyen2021optimal, pmlr-v151-takezawa22a, fan2022complexity, bunne2022proximal, bunne2023schrodinger, bonet2023spherical, mahey2023fast, nguyen2023markovian, nguyen2023unbalanced, hua2023curved,  korotin2023neural, korotin2024light, le2024noisy, nguyen2024sliced}, statistics~\citep{mena2019statistical, pmlr-v99-weed19a, liu2022entropy, nguyen2022many, pmlr-v151-wang22f, nietert2022outlier, nietert2023outlier, pham2024scalable}, and biology~\citep{schiebinger2019optimal, tong2020trajectorynet, bunne2023schrodinger, bunne2023learning, korotin2024light}.

A main drawback of OT is that its computational complexity is high, i.e., super cubic with respect to the number of supports of input measures, which hinders its applications in large-scale applications. Recently, several approaches have been proposed in the literature to scale up the OT problem.~\citet{Cuturi-2013-Sinkhorn} proposes to use entropic regularization for OT and leverages the Sinkhorn algorithm to reduce its complexity into quadratic.~\citet{scetbon2021low} propose low-rank approach to further reduce the computation of Sinkhorn algorithm for entropic regularized OT. Another direction is to exploit the local structure on supports of input measures.~\citet{rabin2011wasserstein} propose sliced-Wasserstein (SW) which projects supports of input measures into a randomly one-dimensional space and utilizes the closed-form computation of the univariate OT. However, by relying on the one-dimensional projection, SW limits its capacity to capture the topological structure of input measures, especially when measures are supported in a high-dimensional space.~\citet{LYFC} propose to leverage tree structure which provides more flexibility and degrees of freedom, i.e., choosing a tree rather than a line as in SW, to alleviate the curse of dimensionality in SW. For practical applications, the tree structure might be a restricted requirement. Recently,~\citet{le2022st} propose a variant of OT, namely Sobolev transport (ST), for measures supported on a graph which admits a closed-form expression for fast computation.

Unlike the classic OT where one can easily utilize various prior geometric structures on supports of input measures, e.g., via the ground cost metric, it is nontrivial to use other structures rather than the $L^p$ structure for the ST. Essentially, the definition of ST is based on the Kantorovich duality of $1$-order Wasserstein, but ST considers the graph-based Sobolev  constraint for the critic function. This makes the ST coupled with the $L^p$ functional geometric structure within the graph-based Sobolev space.

One approach to  move beyond the $L^p$ geometry is to employ the Orlicz geometric structure, which is obtained by leveraging a specific class of convex functions. Indeed, Orlicz norm has been utilized to advance several machine learning approaches in the literature. In particular,~\citet{andoni2018subspace} and~\citet{song2019efficient} leverage the Orlicz norm as a loss for the classic linear regression. Orlicz norm loss regression not only generalizes both the least squared regression (i.e., squared $\ell_2$ norm loss) and the least absolute deviation regression (i.e., $\ell_1$ norm loss), but also provides a scale-invariant version of $\mathbb{M}$-estimator function loss (e.g., the Huber, Cauchy, Welsh, Tukey, Geman-McClure functions~\citep[Table 1]{zhang1997parameter}).\footnote{The classic $\mathbb{M}$-estimation functional loss depends on the data scale, i.e., one may get different solutions if the data are rescaled~\citep{andoni2018subspace}.} Moreover,~\citet{andoni2018subspace} illustrate that the Orlicz norm loss regression improves performances of linear regression with $\ell_1$/$\ell_2$ norms on Gaussian/sparse noise respectively. In addition,~\citet{deng2022fast} propose an efficient data structure for the Orlicz norm as its special case, which helps to scale up many machine learning problems including reinforcement learning, kernelized support vector machine, and clustering. Notably, in order to address the challenging polynomial growing nature of the underlying function class in the empirical process for random Fourier features on approximation of high-order kernel derivatives,~\citet{chamakh2020orlicz} leverage the Orlicz metric on the sample distribution to derive a finite-sample deviation bound for a general class of polynomial-growth functions. In particular, \citet{chamakh2020orlicz} propose finite-sample uniform guarantee for random Fourier features which is almost surely convergence to approximate high-order derivatives for arbitrary kernel.

Furthermore, Orlicz metric has been recently explored in the context of OT. In particular,~\citet{lorenz2022orlicz} propose to leverage Orlicz norm as a regularization for OT problem with continuous probability measures. Especially, Orlicz metric is also utilized as a ground cost for OT, which is known as the Orlicz-Wasserstein (OW)~\citep{sturm2011generalized, kell2017interpolation}. Notably,~\citet{GuhaHN23} and~\citet{altschuler2023faster} have shown that OW possesses unique implicit essences, which may not exist in the classic OT with $L^p$ structure cost, and help to advance certain results in machine learning. In particular,~\citet{GuhaHN23} propose to consider OW metric for studying the Bayesian contraction convergence behavior of parameters arising from hierarchical Bayesian nonparametric models. OW metric helps to alleviate a number of raised concerns caused from the usage of the classic OT with Euclidean cost for quantifying the rates of parameter convergence within infinite Gaussian mixtures to significantly improve the contraction rate. Additionally,~\citet{altschuler2023faster} propose to also leverage OW as a metric shift for R\'enyi divergence to develop novel differential-privacy-inspired techniques to overcome longstanding challenges for proving fast convergence of hypocoercive differential equations. In spite of these achievements, OW  brings up a difficult challenge on its computation. Precisely, OW is a two-level optimization problem: one level is for the transportation plan as in the classic OT problem, and another level is for an extra positive scalar within the Orlicz metric structure.

In this work, we focus on OT problem for probability measures supported on a graph metric space~\citep{le2022st}. On one hand, ST efficiently exploits the graph structure to yield a closed-form for a fast computation. However, ST is essentially coupled with the $L^p$ geometric structure, and it is therefore nontrivial to utilize ST with other prior structures. On the other hand, OW metric owns unique implicit essences to improve certain machine learning problems, but its optimization formulation is challenging for computation. Our goal is to alleviate these issues by adopting Orlicz geometry to generalize the ST for  measures on a graph. This allows us to propose the generalized Sobolev transport (GST) which  inherit the merits from both ST and OW. 

%%%%%%%%%%%%%%%%%%%%%%%%%%%%%%%%%%%%%%%%%%%%
\textbf{Contribution.} In summary, our contributions are three-fold:

\begin{itemize}
\item We leverage a certain class of convex functions corresponding to Orlicz geometric structure, and propose the GST for probability measures supported on a graph metric space. We show that GST can be computed  by simply solving a univariate optimization problem.

\item We demonstrate that ST is a special case of the proposed GST. Additionally, GST utilizes the Orlicz geometric structure in the same sense as the OW for OT problem. We further draw a connection between GST and OW.

\item We empirically illustrate that GST is more computationally efficient than OW. We also show some preliminary evidences on the advantages of GST for document classification and for several tasks in topological data analysis~(TDA).
\end{itemize}

\textbf{Organization.} We briefly review related notations used in the development of our proposals in \S\ref{sec:preliminary}. In \S\ref{sec:GST}, we describe our proposed approach, the generalized Sobolev transport (GST). We show that GST is a metric, and then derive its properties and draw its connection to ST and OW in \S\ref{sec:properties_GST}. In \S\ref{sec:experiments}, we empirically illustrate the computational advantages of GST over OW, and show preliminary improvements of GST on document classification and TDA. In \S\ref{sec:conclusion}, we give concluding remarks. Additionally, we defer the proofs of key theoretical results and additional materials to the Appendices. We have also released code for our proposed approach.\footnote{The code repository is on \url{https://github.com/lttam/Generalized-Sobolev-Transport}}

%%%%%%%%%%%%%%%%%%%%%%%%%%%%%%%%%%%%%%%%%%%%
%%%%%%%%%%%%%%%%%%%%%%%%%%%%%%%%%%%%%%%%%%%%
\section{Preliminaries}\label{sec:preliminary}

In this section, we introduce the notations and give a brief review about graphs together with the Sobolev transport (ST) for measures on a graph, as well as the Orlicz geometric structure and the Orlicz-Wasserstein (OW).

%%%%%%%%%%%%%%%%%%%%%%%%%%%%%%%%%%%%%%%%%%%%
\subsection{Graph and functions on graph}\label{subsec:graph}

We describe the graph setting for measures, and functions on graph. 

%%%%%%%%%%%%%%%%%%%%%%%%%%%%%%%%%%%%%%%%%%%%
\textbf{Graph.} We use the same graph setting as in~\citep{le2022st}. Specifically, let $V$ and $E$ be respectively the sets of nodes and edges. We consider a connected, undirected, and physical\footnote{In the sense that $V$ is a subset of Euclidean space $\R^n$, and each edge $e \in E$ is the standard line segment in $\R^n$ connecting the two vertices  of the edge $e$.} graph $\G = (V,E)$ with positive edge lengths $\{w_e\}_{e\in E}$. Following the convention in~\citep{le2022st} for continuous graph setting, $\G$ is regarded as the set of all nodes in $V$ together with all points forming the edges in $E$. Also,  $\G$ is equipped with the graph metric $d_{\G}(x,y)$  which equals to the length of the shortest path in $\G$ between $x$ and $y$. Additionally, we assume that there exists a fixed root node $z_0 \in V$ such that the shortest path connecting $z_0$ and $x$ is unique for any $x \in \G$, i.e., the uniqueness property of the shortest paths~\citep{le2022st}. 

Let $[x, z]$ denote the shortest path connecting $x$ and $z$ in $\G$. For $x \in \G$, edge $e \in E$, define the sets $\Lambda(x)$, $\gamma_e$ as follow: 
\begin{eqnarray}\label{sub-graph}
 && \Lambda(x) := \big\{y\in \G: \, x\in [z_0,y]\big\}, \nonumber \\
 && \gamma_e := \big\{y\in \G: \, e\subset  [z_0,y]\big\}.
\end{eqnarray}
Denote $\calP(\G)$ (resp.$\,\calP(\G \times\G)$) as the set of all nonnegative Borel measures on $\G$
(resp.$\,\G\times\G$) with a finite mass.

%%%%%%%%%%%%%%%%%%%%%%%%%%%%%%%%%%%%%%%%%%%%
\textbf{Functions on graph.} By a continuous function $f$ on $\G$, we mean that  $f: \G\to \R$ is continuous w.r.t.~the topology on $\G$ induced by the Euclidean distance. Henceforth, $C(\G)$ denotes the set of all continuous functions on $\G$.
Similar notation is used for continuous functions on $\G \times \G$. 

Given a scalar $b>0$, then a function $f:\G\to\R$ is called $b$-Lipschitz w.r.t.~the graph metric $d_\G$ if 
\[
|f(x) - f(y)|\leq b \, d_\G(x,y), \,\, \forall x, y \in \G.
\]

%%%%%%%%%%%%%%%%%%%%%%%%%%%%%%%%%%%%%%%%%%%%
\subsection{Sobolev transport (ST)}

For probability measures on a graph, \citet{le2022st} propose the ST which is a scalable variant of OT. More specifically, \citet{le2022st} leverage the graph structure and propose ST by building upon the dual form of the $1$-order Wasserstein distance, but considering its Lipschitz constraint for the critic function within the graph-based Sobolev space. 

In particular, let $\omega$ be a nonnegative Borel measure on $\G$. Given $1\leq p\leq \infty$, and let  $p'$ be its conjugate, i.e., the number $p'\in [1,\infty]$ satisfying $\frac1p +\frac{1}{p'}=1$. For $\mu, \nu \in \calP(\G)$, the $p$-order Sobolev transport (ST)~\citep[Definition 3.2]{le2022st} is defined as 
\begin{equation} \label{eq:distance}
\hspace{-0.4em} \calS_p(\mu,\nu ) \! \coloneqq \! \left\{
\begin{array}{cl}
\hspace{-0.8em} \sup \Big[\int_\G f(x) \mu(\mathrm{d}x) - \int_\G f(x) \nu(\mathrm{d}x)\Big] \\
\hspace{-0.5em} \mathrm{s.t.} \, f \hspace{-0.2em} \in W^{1,p'} \hspace{-0.3em} (\G, \omega),  \, \|f'\|_{L^{p'}\hspace{-0.2em} (\G, \omega)}\leq 1,
\end{array}
\right.
\end{equation}
where we write $f'$ for the generalized graph derivative of $f$, $W^{1,p'} \hspace{-0.3em} (\G, \omega)$ for the graph-based Sobolev space on $\G$, and $L^{p'}\hspace{-0.2em} (\G, \omega)$ for the $L^p$ functional space on $\G$.\footnote{See Appendix~\S\ref{appsec:Review_SobolevTransport} for a review on the graph-based Sobolev space, the generalized graph derivative function, and the $L^p$ functional space.} Notably, the ST yields a closed-form expression for a fast computation~\citep[Proposition 3.5]{le2022st}.  

Additionally, when graph $\G$ is a tree and $\omega$ is a length measure\footnote{See Definition~\ref{def:measure} in appendix for the length measure.}, the $1$-order Sobolev transport coincides with the $1$-order Wasserstein distance~\citep[Corollary 4.3]{le2022st}.

However, unlike standard OT where one can easily adjust the ground cost following prior structure for applications, it is nontrivial for such adaptation in the ST due to its graph structure and the generalized graph derivation within the graph-based Sobolev space.

%%%%%%%%%%%%%%%%%%%%%%%%%%%%%%%%%%%%%%%%%%%%
\subsection{Orlicz functional space and Orlicz-Wasserstein}\label{subsec:Nfunctions_OW}

We describe a family of convex functions which we leverage to generalize the ST, and briefly review the Orlicz functional geometric structure, as well as the OW which utilizes Orlicz norm as its ground cost. 

%%%%%%%%%%%%%%%%%%%%%%%%%%%%%%%%%%%%%%%%%%%%
\textbf{A family of convex functions.} We consider the collection  of  $N$-functions~\citep[\S8.2]{adams2003sobolev} which are special convex functions on $\R_+$. Hereafter, a strictly increasing and   convex function $\Phi: [0, \infty)\to [0, \infty)$ is called an $N$-function if  $\lim_{t \to 0} \frac{\Phi(t)}{t} = 0$ and $\lim_{t \to +\infty} \frac{\Phi(t)}{t} = +\infty$.

%\textbf{A family of convex functions.} We consider a collection $\mathbb{F}$ of convex function $\Phi : \R_+ \to \R_+$ which is continuous, monotonically increasing, and such that $\Phi(0) = 0$, similar to~\citep{kell2017interpolation, andoni2018subspace}.

\textbf{Examples.} Some popular examples for $N$-functions are (i) $\Phi(t) = t^p$ with $1 < p < \infty$; (ii) $\Phi(t) = \exp(t) - t- 1$; (iii) $\Phi(t) = \exp(t^p) - 1$ with $1 < p < \infty$; and (iv) $\Phi(t) = (1+t) \log(1+t) - t$~\citep[\S8.2]{adams2003sobolev}.

%\tam{Using Huber function as $\Phi$ for Orlicz norm~\citep{andoni2018subspace} -- normalized Huber function?}

%%%%%%%%%%%%%%%%%%%%%%%%%%%%%%%%%%%%%%%%%%%%
\textbf{Orlicz functional space.} Given an  $N$-function $\Phi$ and  
a nonnegative Borel measure $\omega$ on $\G$, let $L_{\Phi}(\G, \omega)$ be the linear hull of the set of all Borel measurable functions $f: \G \to \R$ satisfying $\int_{\G} \Phi(|f(x)|) \omega(\text{d}x) < \infty$. Then, $L_{\Phi}(\G, \omega)$ is a normed space with the Luxemburg norm being defined by
\begin{equation}\label{eq:Luxemburg_norm}
\hspace{-0.2em}\norm{f}_{L_\Phi} \! \coloneqq \! \inf \left\{t > 0 \mid \int_{\G} \Phi\left(\frac{|f(x)|}{t}\right)\omega(\text{d}x) \le 1 \right\}.
\end{equation}
The infimum in Equation~\eqref{eq:Luxemburg_norm} for $\norm{f}_{L_\Phi}$ is attained~\citep[\S8.9]{adams2003sobolev}.

%%%%%%%%%%%%%%%%%%%%%%%%%%%%%%%%%%%%%%%%%%%%
\textbf{Orlicz-Wasserstein (OW).} Following~\citet[Definition 3.2]{GuhaHN23}, the OW with the $N$-function $\Phi$ for measures $\mu, \nu \in \calP(\G)$ is defined as follows:
\begin{eqnarray}\label{eq:OrliczWasserstein}
W_{\Phi}(\mu, \nu) = \inf_{\pi \in \Pi(\mu, \nu)} \inf \Big[ t > 0 : \hspace{7em}\nonumber \\
\int_{\G \times \G} \Phi\left(\frac{d_{\G}(x, z)}{t}\right) \text{d}\pi(x, z) \le 1\Big],
\end{eqnarray}
where $\Pi(\mu, \nu)$ is the set of all  couplings between $\mu$ and $\nu$. 

In this work, we propose to leverage the collection of $N$-functions to generalize the ST, which can adopt Orlicz geometric structure in the same sense as the OW. Therefore, the proposed generalized Sobolev transport can inherit advantages from both ST and OW.

%This approach allows the resulting transport distance to better deal with noisy data, e.g., leveraging convex function based on Huber function.

%%%%%%%%%%%%%%%%%%%%%%%%%%%%%%%%%%%%%%%%%%%%
%%%%%%%%%%%%%%%%%%%%%%%%%%%%%%%%%%%%%%%%%%%%
\section{Generalized Sobolev Transport (GST)}\label{sec:GST}

%namely the set of $N$-functions (\S\ref{subsec:Nfunctions_OW})

In this section, we examine a special family of convex functions which grows faster than linear. By leveraging it, we generalize the ST in Equation~\eqref{eq:distance} by proposing the \emph{graph-based Orlicz-Sobolev space} for the Lipschitz constraint on the critic function for measures on a graph.

%%%%%%%%%%%%%%%%%%%%%%%%%%%%%%%%%%%%%%%%%%%%
\textbf{Graph-based Orlicz-Sobolev space.}

%%%%%%%%%%%%%%%%%%%%%%%%%%%%%%%%%%%%%%%%%%%%
\begin{definition}[Graph-based Orlicz-Sobolev space] \label{def:OrliczSobolev}
Let $\Phi$ be an  $N$-function and $\omega$ be a nonnegative Borel measure on graph $\G$. A continuous function $f: \G \to \R$ is said to belong to the graph-based Orlicz-Sobolev space $\OrliczSobolevPhi(\G, \omega)$ if there exists a function $h\in L_{\Phi}( \G, \omega) $ satisfying 
\begin{equation}\label{eq:OrliczSobolevFunction}
f(x) -f(z_0) =\int_{[z_0,x]} h(y) \omega(\mathrm{d}y),  \quad \forall x\in \G.
\end{equation}
Such function $h$ is unique in $L_{\Phi}(\G, \omega)$ and is called the generalized graph derivative of $f$ w.r.t.~the measure $\omega$. Henceforth, this generalized graph derivative of $f$ is denoted $f'$.
\end{definition}

We remark that the Orlicz-Sobolev space $\OrliczSobolevPhi(\G, \omega)$  depends on the point $z_0$ and the choice of  the $N$-function $\Phi$. For brevity, we however do not explicitly display these dependencies in its notation. Additionally, the formulation in Equation~\eqref{eq:OrliczSobolevFunction} can be considered as a generalized version of the fundamental theorem of calculus, which defines the generalized graph derivative for a continuous function $f$ at any point $x \in \G$. Moreover, the graph-based Orlicz-Sobolev space may be regarded as a generalized version of the graph-based Sobolev space (analyzed rigorously in \S\ref{sec:properties_GST}).

Unlike the OW which directly utilizes the given $N$-function within the Luxemburg norm of the Orlicz functional space (Equation~\eqref{eq:OrliczWasserstein}), in order to generalize the ST, we also require a notion of the complementary function of the given $N$-function.

%%%%%%%%%%%%%%%%%%%%%%%%%%%%%%%%%%%%%%%%%%%%
\textbf{Complementary function.}
For the given $N$-function $\Phi$, its complementary function $\Psi : \R_+ \to \R_+$~\citep[\S8.3]{adams2003sobolev} is the $N$-function, defined as follows
\begin{equation}\label{eq:complementary_func}
\Psi(t) = \sup \left[at - \Phi(a) \, \mid \, a \ge 0  \right], \quad\mbox{for}\,\,\, t\geq 0.
\end{equation}

\textbf{Examples.} Some popular complementary pairs of $N$-functions~\citep[\S8.3]{adams2003sobolev},~\citep[\S 2.2]{rao1991theory} are: (i) $\Phi(t) = \frac{t^p}{p}$ and $\Psi(t) = \frac{t^q}{q}$ where $q$ is the conjugate of $p$, i.e., $\frac{1}{p} + \frac{1}{q} = 1$ and $1 < p < \infty$, and (ii) $\Phi(t) = \exp(t) - t - 1$ and $\Psi(t) = (1+t)\log(1+t) - t$. Additionally, for the $N$-function $\Phi(t) = \exp(t^p) - 1$ with $1 < p < \infty$, its complementary $N$-function admits an explicit expression, but not simple~\citep[\S 2.2]{rao1991theory}.\footnote{See \S\ref{app:subsec:complementary_func_exp} for rigorous details on this complementary function.} 

Inspired by the ST for measures on a graph~\citep{le2022st}, we exploit the dual formulation of the $1$-order Wasserstein distance to propose the \emph{generalized Sobolev transport} (GST). More precisely, we replace the Lipschitz constraint for the critic function by a constraint involving the graph-based Orlicz-Sobolev space.

%%%%%%%%%%%%%%%%%%%%%%%%%%%%%%%%%%%%%%%%%%%%
\begin{definition}[Generalized Sobolev transport distance on graph]
\label{def:GST_distance}
Let $\Phi$ be an $N$-function and $\omega$ be a nonnegative Borel measure on $\G$. For $\mu, \nu\in \calP(\G)$, we define 
\begin{equation} \notag \label{eq:distance_gst}
%\vspace{-0.8em}
\calGS_{\Phi}(\mu,\nu) \! \coloneqq \! \left\{
\begin{array}{cl}
\hspace{-0.4em} \sup &  \Big| \int_\G f(x) \mu(\mathrm{d}x) - \int_\G f(x) \nu(\mathrm{d}x)  \Big| \\
\hspace{-0.4em} \mathrm{s.t.} & f \in {\OrliczSobolevPsi}(\G, \omega),  \, \|f'\|_{L_{\Psi}}\leq 1,
\end{array}
\right.
\end{equation}
where $\Psi$ is the complementary function of $\Phi$ (see \eqref{eq:complementary_func}).
\end{definition}
The Definition~\ref{def:GST_distance} implies that the GST for probability measures supported on a graph metric space is an instance of the integral probability metric~\citep{muller1997integral}.

%%%%%%%%%%%%%%%%%%%%%%%%%%%%%%%%%%%%%%%%%%%%
\textbf{Computation.} We next show that one can compute 
%the GST 
%$\calGS_{\Phi}(\cdot, \cdot)$ 
$\calGS_{\Phi}$ 
by simply solving a univariate optimization problem.

%%%%%%%%%%%%%%%%%%%%%%%%%%%%%%%%%%%%%%%%%%%%
\begin{theorem}[GST as univariate optimization problem]\label{thrm:GST_1d_optimization}
The generalized Sobolev transport $\calGS_{\Phi}(\mu,\nu)$ in Definition~\ref{def:GST_distance} can be computed as follows:
\begin{align}\label{Amemiya_gst}
\calGS_{\Phi}(\mu,\nu )  =  
\inf_{k > 0} \frac{1}{k}\left( 1 + \int_{\G} \Phi(k \left| h(x) \right|) \omega(\text{d}x) \right),
\end{align}
where $h(x) :=  \mu(\Lambda(x)) -  \nu(\Lambda(x))$ for all $x \in \G$.
\end{theorem}
The proof is placed in Appendix \S\ref{app:sec:thrm:GST_1d_optimization}.

We next derive the discrete case for the GST in Equation~\eqref{Amemiya_gst}.

%%%%%%%%%%%%%%%%%%%%%%%%%%%%%%%%%%%%%%%%%%%%
\begin{corollary}[Discrete case]\label{cor:GST_1d_optimization_discrete}
Assume that $\omega(\{x\}) = 0$ for every $x\in \G$, and suppose that $\mu,\nu\in \calP(\G)$ are supported on nodes in $V$ of graph $\G$.\footnote{We discuss an extension for measures supported in $\G$ in \S\ref{lem:length-measure}.} Then, we have
\begin{equation}\label{equ:GST_1d_optimization_discrete}
\calGS_{\Phi}(\mu,\nu )  =  
\inf_{k > 0} \frac{1}{k}\Big[ 1 + \sum_{e \in E} w_e \Phi(k |  \bar{h}(e)|)  \Big],
\end{equation}
where $\bar{h}(e) := \mu(\gamma_e) - \nu(\gamma_e)$ for every edge $e \in E$.
\end{corollary}
The proof is placed in Appendix~\S\ref{app:subsec:cor:GST_1d_optimization_discrete}.\footnote{\citet[Theorem 13]{rao1991theory} derived the necessary and sufficient conditions to obtain the infimum for problem~\eqref{equ:GST_1d_optimization_discrete}.}

From Corollary~\ref{cor:GST_1d_optimization_discrete}, one only needs to simply solve the univariate optimization problem to compute the GST. 

%%%%%%%%%%%%%%%%%%%%%%%%%%%%%%%%%%%%%%%%%%%%
\begin{remark}[GST for non-physical graph]
Similar to the ST, we have assumed that $\G$ is a physical graph in \S\ref{subsec:graph}. However, Corollary~\ref{cor:GST_1d_optimization_discrete} implies that the GST $\calGS_{\Phi}$ does not depend on the physical assumption when input measures are supported on graph nodes. In particular, it only depends on the graph structure $(V, E)$ and edge weights $w_e$. Thus, we can apply the GST for non-physical graph $\G$.
\end{remark}

\begin{remark}[Complementary pairs of $N$-functions for GST.]
The Definition~\ref{def:GST_distance} for GST with the $N$-function $\Phi$ involves its complementary $N$-function $\Psi$. However, we can reformulate GST as a univariate optimization problem without involving the complementary function $\Psi$ as in Equation~\eqref{Amemiya_gst}. Note that in order to obtain the univariate optimization formulation, the complementary function $\Psi$ is finite-valued, which is satisfied by any $N$-function $\Phi$, i.e., growing faster than linear (see \eqref{eq:complementary_func}).      
\end{remark}

\textbf{Preprocessing.} For the computation of GST in Equation~\eqref{equ:GST_1d_optimization_discrete}, observe that set $\gamma_e$ (see Equation~\eqref{sub-graph}) can be precomputed for all edge $e$ in $\G$. This preprocessing step only involves the graph $\G$ itself, and is not related to input measures. Moreover, we only need to precompute it once, regardless the number of input pairs of probability measures
%, which we will 
to be measured by the GST. In particular, we recompute the shortest paths from the root node $z_0$ to all other input supports (or vertices) by Dijkstra algorithm with the complexity $\mathcal{O}(|E| + |V| \log{|V|})$, where $|\cdot|$ is the cardinality of a set. 

%edge $e$ belongs to the shortest path from the root node $z_0$ to $x$, i.e., 
\begin{remark}[Sparsity in Problem~\eqref{equ:GST_1d_optimization_discrete}] Let $\text{supp}(\mu)$ be the set of supports of $\mu$. Observe that for any support $x \in \text{supp}(\mu)$, its mass is accumulated into $\mu(\gamma_e)$ if and only if $e \subset [z_0, x]$. Let define set $E_{\mu, \nu} \subset E$ as
\[
E_{\mu, \nu} \hspace{-0.2em}:=\hspace{-0.2em} \left\{e \! \in \! E \mid \exists z \! \in \! (\text{supp}(\mu) \cup \text{supp}(\nu)), e \subset [z_0, z] \right\}\!.
\]
Then in \eqref{equ:GST_1d_optimization_discrete}, we in fact only need to take the summation over all edges $e \in E_{\mu, \nu}$, i.e., removing all edges $e \in E \setminus E_{\mu, \nu}$.
\end{remark}

%%%%%%%%%%%%%%%%%%%%%%%%%%%%%%%%%%%%%%%%%%%%
%%%%%%%%%%%%%%%%%%%%%%%%%%%%%%%%%%%%%%%%%%%%
\section{Properties of the GST}\label{sec:properties_GST}

In this section, we derive the metric property for the GST and establish a relationship for the GST with different $N$-functions. Additionally, we draw connections of the GST to the ST, the OW, and OT.

%%%%%%%%%%%%%%%%%%%%%%%%%%%%%%%%%%%%%%%%%%%%
\begin{theorem}[Metrization]\label{thrm:metrization}
The generalized Sobolev transport $\calGS_{\Phi}(\mu,\nu)$ is a metric on the space $\calP(\G)$.
\end{theorem}
The proof is placed in Appendix~\S\ref{app:subsec:thrm:metrization}.

% using \calP(\G) (instead of \calM(\G)) for the set of all nonegative Borel measures on \G

The GST is monotone with respect to the $N$-function $\Phi$
as shown in the next result.
Consequently, it may enclose a stronger notion of metrics than the ST for comparing measures on a graph.
\begin{proposition}[GST with different $N$-functions]\label{prop:strong_metric}
For any two $N$-functions $\Phi_1, \Phi_2$ satisfying  $\Phi_1(t) \le \Phi_2(t)$ for all $t \in \R_+$,  we have
\[
\calGS_{\Phi_1}(\mu,\nu ) \le \calGS_{\Phi_2}(\mu,\nu )
\,\,\,\mbox{for every}\,\,\, \mu, \nu \in \calP(\G).
\]
\end{proposition}
The proof is placed in Appendix~\S\ref{app:subsec:prop:strong_metric}.

%%%%%%%%%%%%%%%%%%%%%%%%%%%%%%%%%%%%%%%%%%%%
\textbf{Connection with Sobolev transport.} We next show the connection between the graph-based Orlicz-Sobolev space and the graph-based Sobolev space~\citep{le2022st}.

\begin{proposition}\label{prop:Connection_OS_Sobolev}
Let $1 < p < \infty$ and $\omega$ be a nonnegative Borel measure  on graph $\G$.
By taking $\Phi(t) := t^p$ for the $N$-function, then the associated graph-based Sobolev-Orlicz space is the same as the graph-based Sobolev space of order $p$, denoted as $W^{1, p}(\G, \omega)$.\footnote{See Definition~\ref{def:Sobolev} for the graph-based Sobolev space.} That is,
\[
\OrliczSobolevPhi(\G, \omega) = W^{1, p}(\G, \omega).
\]
\end{proposition}

The proof is placed in Appendix~\S\ref{app:subsec:prop:Connection_OS_Sobolev}.

By leveraging the result in Proposition~\ref{prop:Connection_OS_Sobolev}, we next derive the closed-form expression for the univariate optimization problem in~\eqref{Amemiya_gst} for the GST with certain $N$-functions. This allows us to draw its connection to the ST~\citep{le2022st}.

\begin{proposition}[Connection between GST and ST]\label{prop:Connection_GST_ST}
Let $\Phi(t) := \frac{(p-1)^{p-1}}{p^p} t^p$ with $1 < p < \infty$, and  let $\omega$ be a nonnegative Borel measure  on graph $\G$. Then, the GST admits the following closed-form expression:
\begin{equation}\label{eq:GST_ST}
\calGS_{\Phi}(\mu,\nu ) = \left( \int_{\G} \left| \mu(\Lambda(x)) -  \nu(\Lambda(x)) \right|^p \omega(\text{d}x) \right)^{\frac{1}{p}},
\end{equation}
for all measures $\mu, \nu \in \calP(\G)$.
Consequently, we have 
\[
\calGS_{\Phi}(\mu,\nu ) = \calS_p(\mu,\nu ).
\]
\end{proposition}
The proof is placed in Appendix~\S\ref{app:subsec:prop:Connection_GST_ST}.

%%%%%%%%%%%%%%%%%%%%%%%%%%%%%%%%%%%%%%%%%%%%
\textbf{Connection with Orlicz-Wasserstein and OT.} In the special case $\Phi(t) = t^p$ with $p \ge 1$, the OW coincides with the $p$-order Wasserstein~\citep{GuhaHN23}.\footnote{The $p$-order Wasserstein is reviewed in \S\ref{def:pWasserstein}. Although $\Phi(t) = t$ is not an $N$-function due to its linear growth, it can be regarded as the limit $p \to 1^+$ for the function $\Phi(t) = t^p$. In the similar way,~\citet{andoni2018subspace} leverages the Huber function with linear growth for $\Phi(t)$ for Orlicz loss regression in applications.} It is worth noting that OW in Equation~\eqref{eq:OrliczWasserstein} is a two-level optimization problem involving the transportation plan and  the positive scalar $t$. Thus, it is more challenging to compute the OW than the classic OT (e.g., with squared Euclidean cost).

\begin{remark}\label{rm:OW-TW}
Following~\citet{GuhaHN23},  the OW is equal to the $1$-order Wasserstein if we take  $\Phi(t) = t^p$  and then take the limit $p \to 1^+$. In the similar way and by taking the limit $p \to 1^+$, the closed-form GST in Equation~\eqref{eq:GST_ST} is equal to the $1$-order ST.\footnote{Note that $\lim_{p \to 1^+} \frac{(p-1)^{p-1}}{p^p} t^p = t$, see~\S\ref{app:subsec:rm:OW-TW} for the proof.} With an additional assumption that graph $\G$ is a tree, it follows from \citet{le2022st} that the $1$-order ST is in turn equal to the $1$-order Wasserstein. Thus, the GST coincides with the OW when $\Phi(t) = t$ and the graph $\G$ is a tree. 
\end{remark}

Due to Remark~\ref{rm:OW-TW}, the GST can be regarded as a variant of the OW. We summarize it in the following Proposition~\ref{prop:GST_OW}.
\begin{proposition}[Connection between GST and OW]\label{prop:GST_OW}
When $\Phi(t) = t$ and the graph $\G$ is a tree, the generalized Sobolev transport and Orlicz-Wasserstein coincide, i.e.,
\[
\calGS_{\Phi}(\mu,\nu ) = W_{\Phi}(\mu, \nu),
\]
for all measures $\mu, \nu \in \calP(\G)$.
\end{proposition}
Notice that our obtained results show that one only needs to solve a univariate optimization problem to compute the GST. On the other hand, it is much more challenging to solve the two-level optimization problem \eqref{eq:OrliczWasserstein} for the OW. Notably, a recent realization is that solving exactly OT problems leads to overfitting~\citep[\S8.4]{peyre2019computational}. Therefore, it would be self-defeating to give excessive efforts to optimize the OT problems since it would lead to overfitting within the computation of the OT problems themselves. Consequently, GST can be regarded as a regularization approach for OW.

% ==========================================
% Orlicz function used in the literature
% ==========================================
\begin{comment}
Note that we will need the complementary function of the given Orlicz function for the computation of the generalized Sobolev transport.

\begin{itemize}
\item \citet{altschuler2023faster} used the sub-Gaussian Orlicz function $\Phi(t) = \exp(t^2) - 1$~\citep[\S2.7.1]{vershynin2018high}

\item \citet{GuhaHN23} used $\Phi(t) = \exp(t/\beta) - 1$ with $\beta = 1.1$.

\item \citet{lorenz2022orlicz} used $\Phi(t) = t^2/2$ and a variant of $\Phi(t) = t \log(t)$ (see $\Phi(t) = (t+1)\log(t+1) - t$ for more popular Orlicz function.)

\item \citet{chamakh2020orlicz} used $\Phi(t) = \exp(t^{\alpha}) - 1$.

\item Huber-based function???

\end{itemize}
\end{comment}
% ==========================================
% ==========================================

%%%%%%%%%%%%%%%%%%%%%%%%%%%%%%%%%%%%%%%%%%%%
%%%%%%%%%%%%%%%%%%%%%%%%%%%%%%%%%%%%%%%%%%%%
\section{Experiments}\label{sec:experiments}

In this section, we illustrate that the computation of OW for general $N$-function $\Phi$ is very expensive, and the GST is several-order faster than OW. Then, we evaluate the GST and show preliminary evidences on its advantages for document classification and for several tasks in TDA. 

%($M=10^3$)
% with different $\Phi(\cdot)$ functions on $10^4$ pairs of measures supported 
\begin{figure}[h]
%\begin{wrapfigure}{r}{0.22\textwidth}
  \vspace{3pt}
  \begin{center}
    \includegraphics[width=0.3\textwidth]{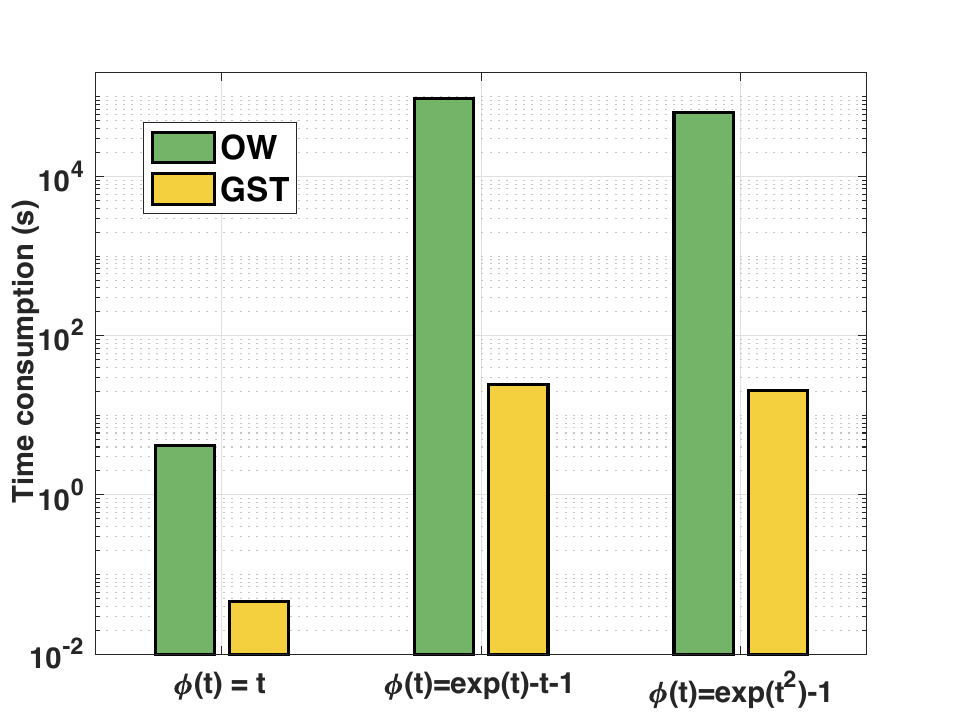}
  \end{center}
%  \vspace{-14pt}
  \vspace{-6pt}
  \caption{Time consumption for GST and OW on $\G_{\text{Log}}$.}
  \label{fg:Time_GST_OW_10K_LLE}
 %\vspace{-6pt}
\end{figure}
%\end{wrapfigure}

%($M=10^3$)
% with different $\Phi(\cdot)$ functions on $10^4$ pairs of measures supported 
\begin{figure}[h]
%\begin{wrapfigure}{r}{0.22\textwidth}
%  \vspace{-6pt}
  \begin{center}
    \includegraphics[width=0.3\textwidth]{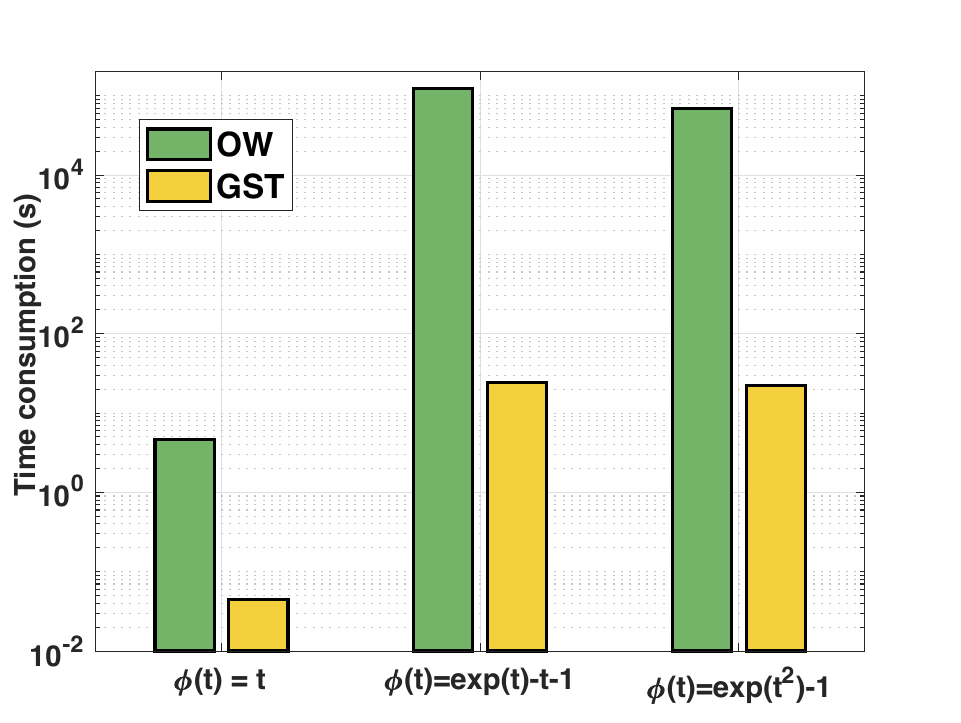}
  \end{center}
%  \vspace{-14pt}
  \vspace{-6pt}
  \caption{Time consumption for GST and OW on $\G_{\text{Sqrt}}$.}
  \label{fg:Time_GST_OW_10K_SLE}
% \vspace{-6pt}
\end{figure}
\begin{figure*}[ht]
%\begin{wrapfigure}{r}{0.22\textwidth}
%  \vspace{-6pt}
  \begin{center}
    \includegraphics[width=0.65\textwidth]{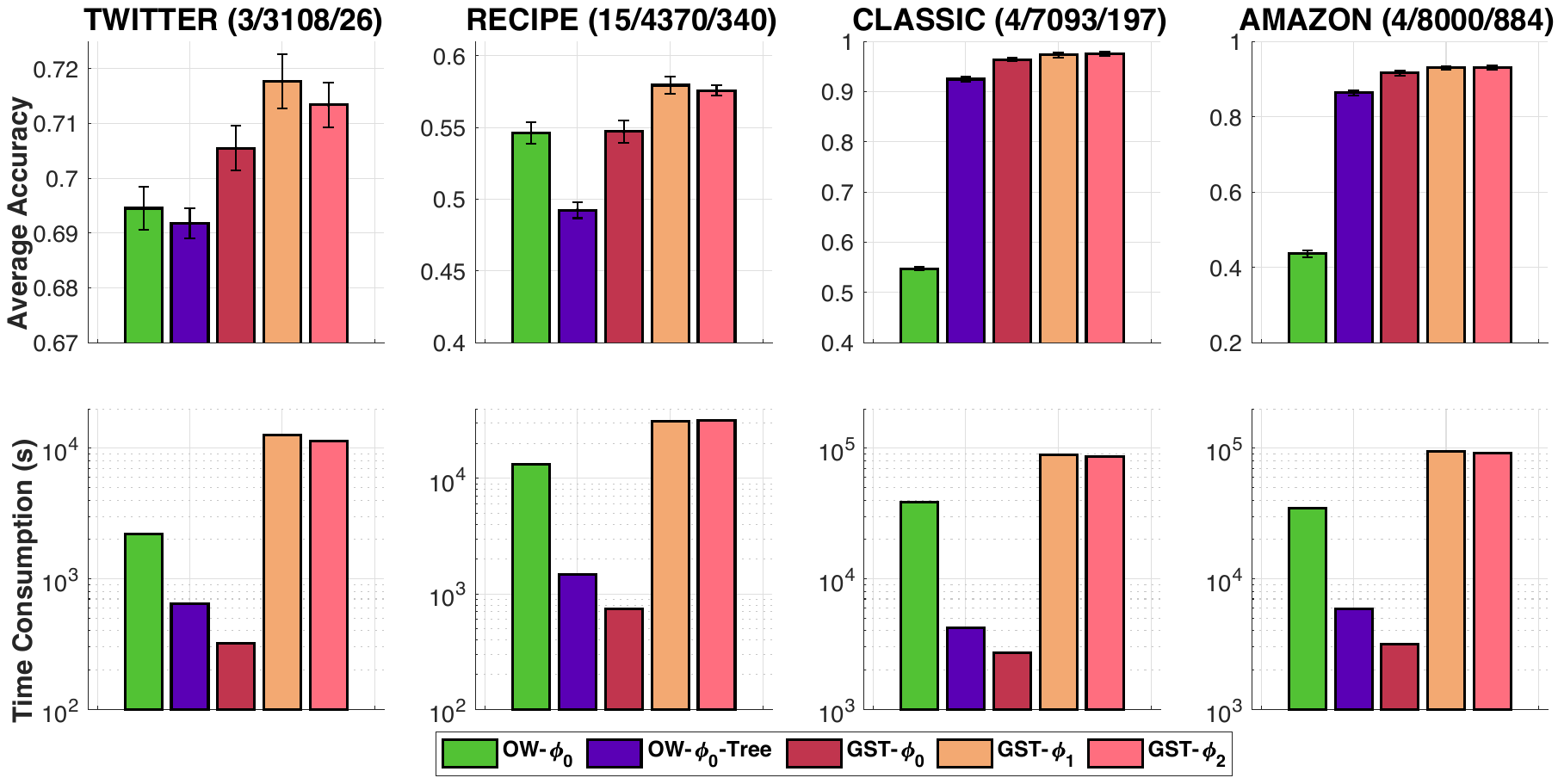}
  \end{center}
  \vspace{-6pt}
  \caption{Document classification on graph $\G_{\text{Log}}$. For each dataset, the numbers in the parenthesis are respectively the number of classes; the number of documents; and the maximum number of unique words for each document.}
  \label{fg:DOC_LLE_10K}
% \vspace{-6pt}
\end{figure*}

%\end{wrapfigure}
\textbf{Document classification.} We consider $4$ traditional document datasets: \texttt{TWITTER, RECIPE, CLASSIC, AMAZON}. The characteristic properties of these datasets are summarized in Figure~\ref{fg:DOC_LLE_10K}. Following~\citet{le2022st}, we use word embedding, e.g., word2vec~\citep{mikolov2013distributed} pretrained on Google News for documents, to map words into vectors in~$\R^{300}$. Additionally, we remove SMART stop words~\citep{salton1988term}, and words in documents which are not in the pretrained word2vec. We then regard each document as a probability measure by considering each word in the document as its support in~$\R^{300}$, and using word frequency as the mass on that corresponding word.

%le2018persistence,
\textbf{TDA.} We consider two tasks: orbit recognition on the synthesis \texttt{Orbit} dataset~\citep{adams2017persistence}, and object shape classification on \texttt{MPEG7} dataset~\citep{latecki2000shape} as in \citet{le2022st}. The characteristic properties of these datasets are summarized in Figure~\ref{fg:TDA_LLE_10K1K}. We utilize the persistence diagrams (PD) to represent objects of interest for these tasks. Particularly, PD is a multiset of points in $\R^2$, where each point summarizes a life span (i.e., birth and death time) of a particular topological feature (e.g., connected component, ring, or cavity), extracted by algebraic topology methods (e.g., persistence homology)~\citep{edelsbrunner2008persistent}. We consider each PD as a probability measure of the $2$-dimensional topological feature data points with a uniform mass.

% ($M=10^4$)
%%%%%%%%%%%%%%%%%%%%%%%%%%%%%%%%%%%%
% 10K
%\end{wrapfigure}

% ($M=10^4$)
\begin{figure*}[ht]
%\begin{wrapfigure}{r}{0.22\textwidth}
%  \vspace{-6pt}
  \begin{center}
    \includegraphics[width=0.65\textwidth]{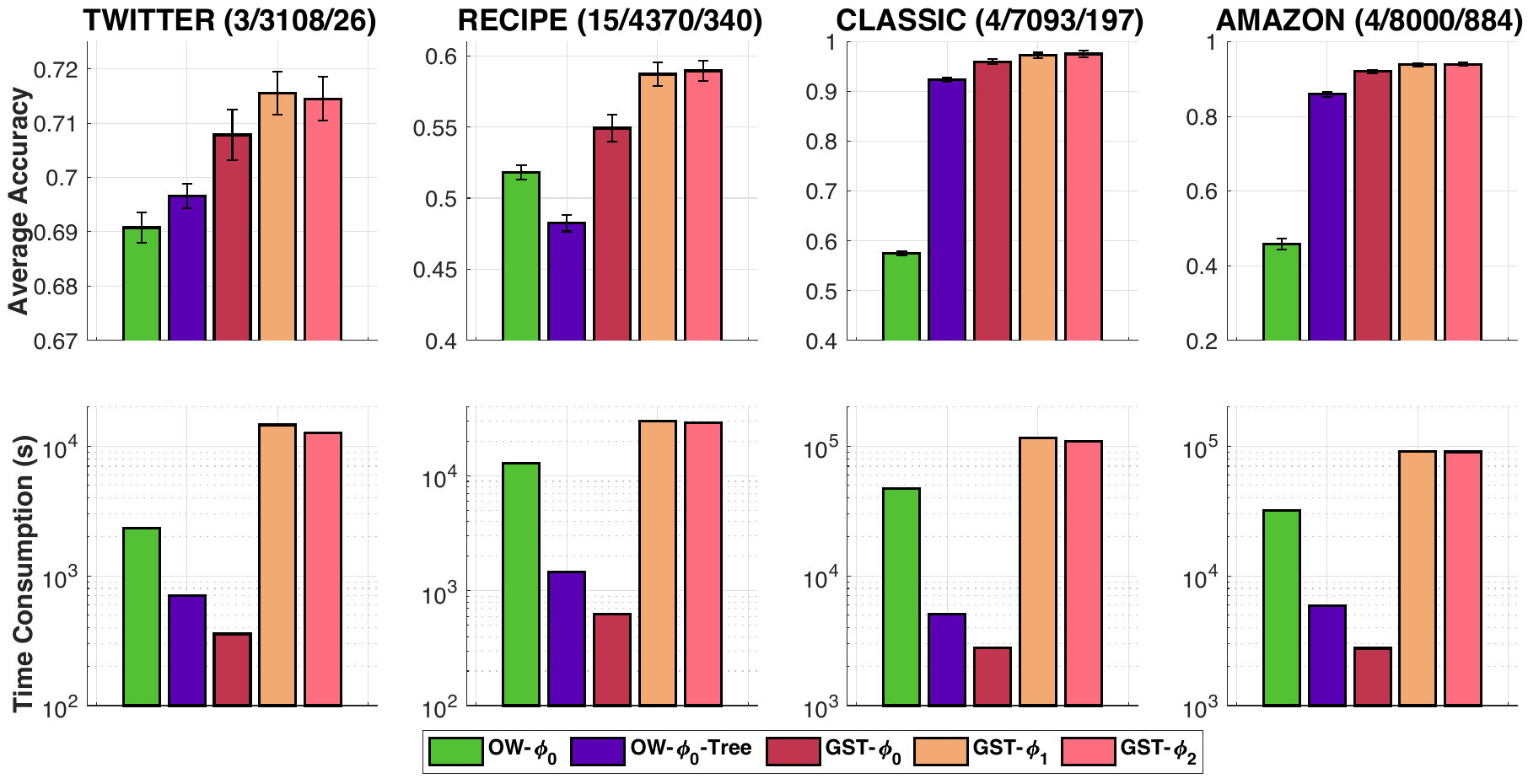}
  \end{center}
  \vspace{-6pt}
  \caption{Document classification on graph $\G_{\text{Sqrt}}$.}
  \label{fg:DOC_SLE_10K}
 \vspace{-2pt}
\end{figure*}
%\end{wrapfigure}

\textbf{Graph.} We employ the graphs $\G_{\text{Log}}$ and $\G_{\text{Sqrt}}$~\citep[\S5]{le2022st} for our simulations, which empirically satisfy the assumptions in \S\ref{sec:preliminary}.\footnote{See Appendix~\S\ref{app:subsec:discussion} for a review and further discussions.} Additionally, we consider $M\!=\! 10^2, 10^3, 10^4$ as the number of nodes for these graphs.

\textbf{$N$-function.} We consider popular $N$-functions $\Phi$ for the GST and the OW: $\Phi_1(t) = \exp(t) - t - 1$, and $\Phi_2(t) = \exp(t^2) - 1$. We also examine the limit case: $\Phi_0(t) = t$.

\textbf{Optimization algorithm.} For general $N$-functions, we use~\citet[Algorithm 1]{GuhaHN23} to compute OW, and a second-order method, e.g., fmincon Trust Region Reflective solver in MATLAB, for solving the \emph{univariate} optimization problem to compute the GST. 

\begin{table}[]
%\begin{wraptable}{r}{0.2\textwidth}
\vspace{-6pt}
\caption{The number of pairs on datasets for SVM.}
\label{tb:numpairs}
    \centering
\begin{tabular}{|l|c|}%{|l|l|}
\hline
Datasets & \#pairs \\ \hline
\texttt{TWITTER}  & 4394432 \\ \hline
\texttt{RECIPE}   & 8687560      \\ \hline
\texttt{CLASSIC}  & 22890777       \\ \hline
\texttt{AMAZON}   & 29117200      \\ \hline
\texttt{Orbit}    & 11373250   \\ \hline
\texttt{MPEG7}    & 18130     \\ \hline
\end{tabular}
\vspace{-14pt}
%\end{wraptable}
\end{table}

%\citet{Cuturi-2013-Sinkhorn, ref:le2019tree}
\textbf{Classification.} We use support vector machine (SVM) for both document classification and the tasks in TDA. We employ kernel $\exp(-\bar{t} \bar{d}(\cdot, \cdot))$, where $\bar{d}$ is a distance (e.g., GST, OW) for probability measures supported on a graph, and $\bar{t} > 0$. Following~\citet{Cuturi-2013-Sinkhorn}, we regularize for the Gram matrices by adding a sufficiently large diagonal term for SVM with indefinite kernels. We use 1-vs-1 strategy to carry out SVM for multi-class classification, e.g., with Libsvm.\footnote{https://www.csie.ntu.edu.tw/$\sim$cjlin/libsvm/} For each dataset, we randomly split it into $70\%/30\%$ for training and test with $10$ repeats. We typically choose hyper-parameters via cross validation. For kernel hyperparameter, we choose $1/\bar{t}$ from $\{q_{s}, 2q_{s}, 5q_{s}\}$ with $s = 10, 20, \dotsc, 90$ where $q_s$ is the $s\%$ quantile of a subset of distances observed on a training set. For SVM regularization hyperparameter, we choose it from $\left\{0.01, 0.1, 1, 10\right\}$. For the root node $z_0$ in graph $\G$, we choose it from a random $10$-root-node subset of $V$ in $\G$. For document classification and TDA, reported time consumption includes preprocessing procedures, e.g., computing shortest paths for GST and OW.

We also summarize the number of pairs which we need to compute distances in training and test with kernel SVM for each run in Table~\ref{tb:numpairs}, e.g., more than $29$ million pairs of measures on \texttt{AMAZON} dataset. 

%%%%%%%%%%%%%%%%%%%%%%%%%%%%%%%%%%%%%%%%%%%%
\subsection{Computation}\label{subsec:computation}

We compare the time consumption of GST and OW with popular $N$-functions $\Phi_1, \Phi_2$, and the limit case $\Phi_0$.\footnote{Notice that GST-$\Phi_0$ is equal to $1$-order Sobolev transport (see Remark~\ref{rm:OW-TW}), and OW-$\Phi_0$ is equal to the OT on a graph (i.e., OT with graph metric ground cost).}
%Thus, both GST-$\Phi_0$ and OW-$\Phi_0$ are metric.

\textbf{Setup.} We randomly sample $10^4$ pairs of measures from \texttt{AMAZON} dataset, and consider graphs $\G_{\text{Log}}$ and $\G_{\text{Sqrt}}$ with the number of nodes $M = 10^3$.

\textbf{Results and discussions.} We illustrate the time consumption on $\G_{\text{Log}}$ and $\G_{\text{Sqrt}}$ in Figures~\ref{fg:Time_GST_OW_10K_LLE} and \ref{fg:Time_GST_OW_10K_SLE} respectively. GST is several-order faster than OW. More concretely, GST is $100\times, 5000\times, 3000\times$ faster than OW for $\Phi_0, \Phi_1, \Phi_2$ respectively. Especially, for $\Phi_1$ and $\Phi_2$, GST takes less than \emph{$25$ seconds}, but OW takes at least \emph{$17$ hours}, and up to \emph{$33$ hours}. 

Notice that the OW with $\Phi_0$ is equal to the OT with graph metric $d_{\G}$ as its ground cost. Additionally, GST with $\Phi_0$ yields a closed-form expression for a fast computation, following the Proposition~\ref{prop:Connection_GST_ST} and Remark~\ref{rm:OW-TW}. Therefore, the computation of GST and OW with $\Phi_0$ is more efficient than with $\Phi_1, \Phi_2$.

%%%%%%%%%%%%%%%%%%%%%%%%%%%%%%%%%%%%%%%%%%%%
\subsection{Document classification}

\textbf{Set up.} We evaluate GST with $\Phi_0, \Phi_1, \Phi_2$ as in \S\ref{subsec:computation} (denoted as GST-$\Phi_i$ for $i = 0, 1, 2$). For OW, we only consider $\Phi_0$ (denoted as OW-$\Phi_0$), but exclude $\Phi_1, \Phi_2$ due to their heavy computations (illustrated and discussed in \S\ref{subsec:computation}). Following~\citet{le2022st} and Proposition~\ref{prop:GST_OW}, we also consider a special case for OW-$\Phi_0$, where we randomly sample a tree from the given graph $\G$ (denoted as OW-$\Phi_0$-Tree). Notice that OW-$\Phi_0$-Tree admits a closed-form expression for a fast computation~\citep{LYFC}. We report results on graphs $\G_{\text{Log}}$ and $\G_{\text{Sqrt}}$ with the number of nodes $M=10^4$. Further empirical results with different values for $M$ are placed in Appendix~\S\ref{app:subsec:further_empirical_results}.

\textbf{Results and discussions.} We illustrate SVM results and the time consumption of kernel matrices on graphs $\G_{\text{Log}}$ and $\G_{\text{Sqrt}}$ in Figures~\ref{fg:DOC_LLE_10K} and~\ref{fg:DOC_SLE_10K} respectively. The performances of GST with all $\Phi$ functions compare favorably to those of OW. Additionally, the computation of GST-$\Phi_0$ is several-order faster than OW-$\Phi_0$. We also reemphasize that it is prohibitively expensive to evaluate OW with $\Phi_1, \Phi_2$ functions on many pairs of measures (See Table~\ref{tb:numpairs} and \S\ref{subsec:computation}). GST-$\Phi_1$ and GST-$\Phi_2$ improve performances of GST-$\Phi_0$, but their computational time is several-order higher (i.e., GST-$\Phi_0$ has a closed-form expression for a fast computation). Thus, it may imply that Orlicz geometric structure in GST may be useful for document classification. Recall that OW-$\Phi_0$-Tree uses a partial information of $\G$, while OW-$\Phi_0$ uses information of the whole $\G$. On the other hand, kernel for OW-$\Phi_0$-Tree is positive definite, while kernel for OW-$\Phi_0$ is indefinite. Performances of OW-$\Phi_0$-Tree are worse than those of OW-$\Phi_0$ in \texttt{TWITTER, RECIPE}, but better than those of OW-$\Phi_0$ in \texttt{CLASSIC, AMAZON}, which agrees with observations in~\citet{le2022st}. 

% ($M=10^4$ for \texttt{Orbit}, and $M=10^3$ for \text{MPEG7})
%%%%%%%%%%%%%%%%%%%%%%%%%%%%%%%%%%
% 10K1K
\begin{figure}[h]
%\begin{wrapfigure}{r}{0.22\textwidth}
%  \vspace{-6pt}
  \begin{center}
    \includegraphics[width=0.35\textwidth]{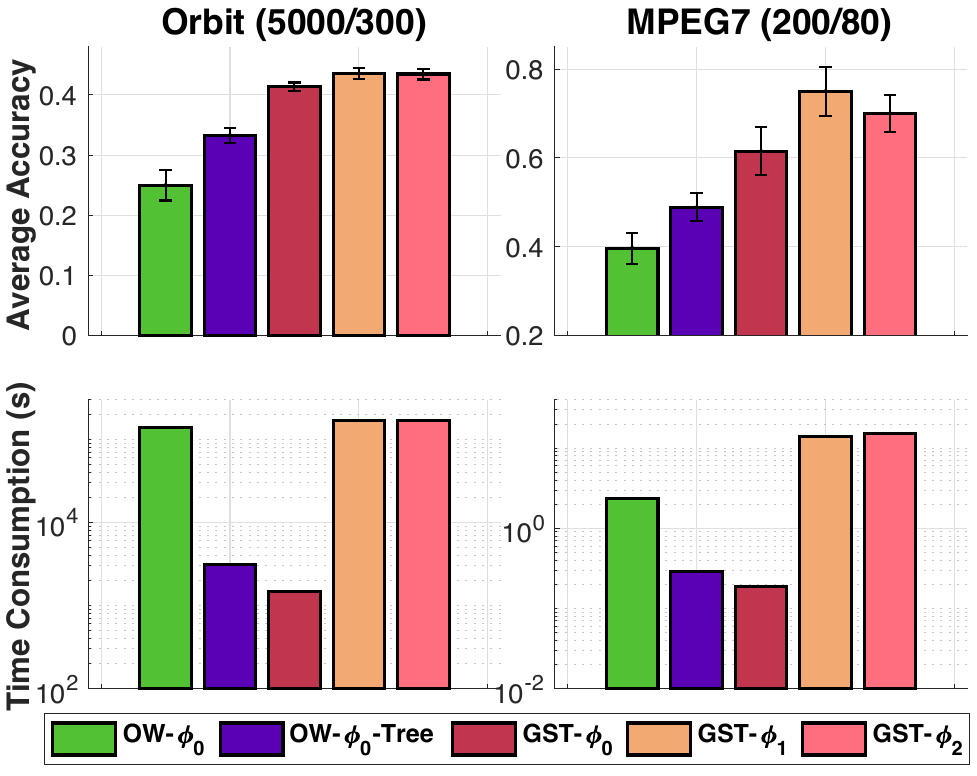}
  \end{center}
  \vspace{-6pt}
  \caption{TDA on graph $\G_{\text{Log}}$. For each dataset, the numbers in the parenthesis are respectively the number of PD; and the maximum number of points in PD.}
  \label{fg:TDA_LLE_10K1K}
% \vspace{-6pt}
\end{figure}
%\end{wrapfigure}

% ($M=10^4$ for \texttt{Orbit}, and $M=10^3$ for \text{MPEG7})
\begin{figure}[h]
%\begin{wrapfigure}{r}{0.22\textwidth}
%  \vspace{-6pt}
  \begin{center}
    \includegraphics[width=0.35\textwidth]{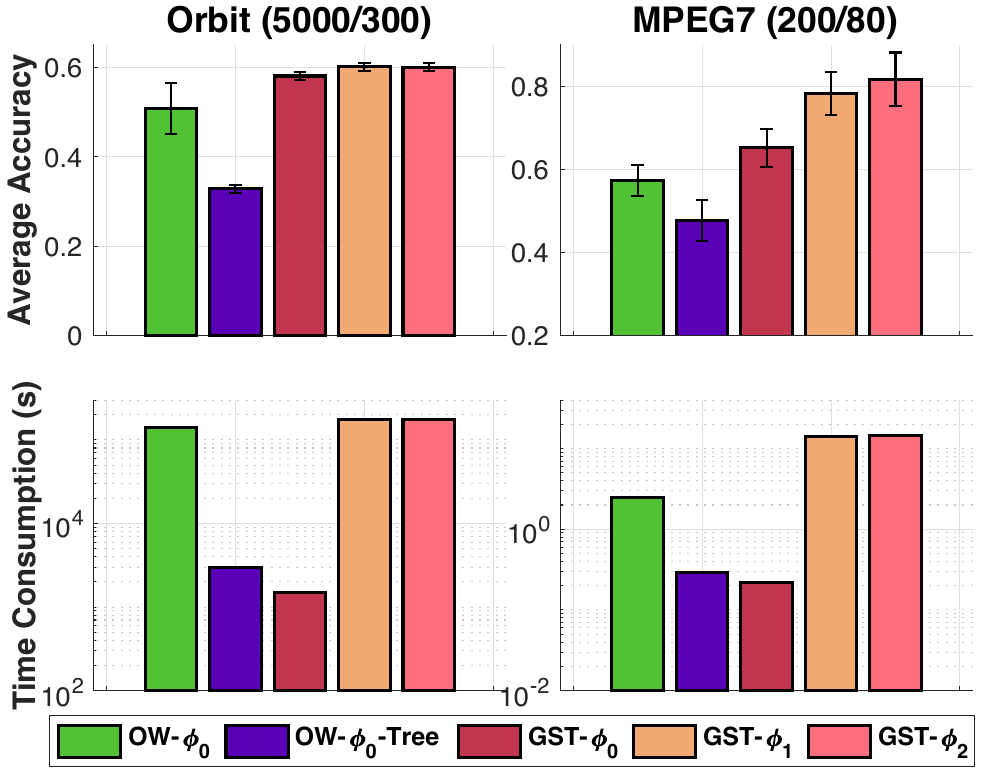}
  \end{center}
  \vspace{-6pt}
  \caption{TDA on graph $\G_{\text{Sqrt}}$.}
  \label{fg:TDA_SLE_10K1K}
% \vspace{-6pt}
\end{figure}
%\end{wrapfigure}
%%%%%%%%%%%%%%%%%%%%%%%%%%%%%%%%%%%%%%%%%%%%
\subsection{Topological Data Analysis}

\textbf{Set up.} We evaluate the same distances as in document classification, i.e., GST-$\Phi_i$ for $i = 0, 1, 2$; OW-$\Phi_0$; and OW-$\Phi_0$-Tree. We report results on graphs $\G_{\text{Log}}$ and $\G_{\text{Sqrt}}$ with $M=10^4$ for \texttt{Orbit} dataset, and $M=10^3$ for \texttt{MPEG7} dataset due to its small size. We also include further empirical results with different values for $M$ in Appendix~\S\ref{app:subsec:further_empirical_results}.

\textbf{Results and discussions.} We illustrate SVM results and the time consumption of kernel matrices on graphs $\G_{\text{Log}}$ and $\G_{\text{Sqrt}}$ in Figures~\ref{fg:TDA_LLE_10K1K} and~\ref{fg:TDA_SLE_10K1K} respectively. We have similar observations as for document classification. GST is several-order faster than OW (for the same $\Phi$ function), and improve performances of OW. Orlicz geometry structure is also helpful for TDA. Although only using a partial information of $\G$, positive definiteness may play an important role for OW-$\Phi_0$-Tree to improve performances of OW-$\Phi_0$ in TDA, also observed in~\citet{le2022st}.

%%%%%%%%%%%%%%%%%%%%%%%%%%%%%%%%%%%%%%%%%%%%
%%%%%%%%%%%%%%%%%%%%%%%%%%%%%%%%%%%%%%%%%%%%
\section{Conclusion}\label{sec:conclusion}

We propose the generalized Sobolev transport (GST) for probability measures supported on a graph metric space.
This is achieved 
by leveraging a special family of convex functions %which grow faster than linear
(i.e., the set of~$N$-functions) to extend the Sobolev transport (ST), which is a scalable variant of OT on a graph. This novel approach enables us to adopt Orlicz geometric structure for GST. Consequently, GST can inherit advantages from both ST and the Orlicz-Wasserstein (OW). %Especially, 
An important feature is  that unlike OW which involves a complex two-level optimization problem, one can simply solve a univariate optimization problem for the GST computation. Thus, GST is a computationally efficient variant for the computationally-demanding OW, which paves a way to use OW in applications. For future work, it is interesting to extend the GST for measures possibly having different total mass (i.e., unbalanced setting), which is essential for applications having noisy supports or outliers in input measures~\citep{frogner2015learning, balaji2020robust, mukherjee2021outlier, fatras2021unbalanced}.

\section*{Acknowledgements}

We thank the area chairs and anonymous reviewers for their comments. KF has been supported in part by Grant-in-Aid for Transformative Research Areas (A) 22H05106. TL gratefully acknowledges the support of JSPS KAKENHI Grant number 23K11243, and Mitsui Knowledge Industry Co., Ltd. grant.

\section*{Impact Statement}

The paper presents a generalized version for the Sobolev transport (ST) which is a scalable variant of optimal transport for probability measures on a graph. By leveraging a special family of convex functions, we can adopt Orlicz geometric structure for the generalized Sobolev transport (GST). Furthermore, GST can be regarded as a computationally efficient variant of the computationally-demanding Orlicz-Wasserstein (OW), which possesses implicit essences to advance certain results in machine learning. Therefore, the proposed GST can inherit the merits from both ST and OW. To our knowledge, there are no foresee potential societal consequences of our work.


%\bibliography{example_paper}
%\bibliographystyle{icml2024}

\balance

%\bibliography{example_paper}
\bibliographystyle{icml2024}

\bibliography{bibEPT21, bibSobolev22}

%%%%%%%%%%%%%%%%%%%%%%%%%%%%%%%%%%%%%%%%%%%%%%%%%%%%%%%%%%%%%%%%%%%%%%%%%%%%%%%
%%%%%%%%%%%%%%%%%%%%%%%%%%%%%%%%%%%%%%%%%%%%%%%%%%%%%%%%%%%%%%%%%%%%%%%%%%%%%%%
% APPENDIX
%%%%%%%%%%%%%%%%%%%%%%%%%%%%%%%%%%%%%%%%%%%%%%%%%%%%%%%%%%%%%%%%%%%%%%%%%%%%%%%
%%%%%%%%%%%%%%%%%%%%%%%%%%%%%%%%%%%%%%%%%%%%%%%%%%%%%%%%%%%%%%%%%%%%%%%%%%%%%%%
\newpage
\appendix
\onecolumn

\begin{center}
{\bf{\Large{Supplement to  ``Generalized Sobolev Transport for Probability Measures on a Graph"}}}
\end{center}

In \S\ref{app:sec:proof} of this appendix, we give the detailed proofs for our theoretical results. Further results and discussions are given in \S\ref{app:sec:further_results_discussions}.

%%%%%%%%%%%%%%%%%%%%%%%%%%%%%%%%%%%%%%%%%%%%
%%%%%%%%%%%%%%%%%%%%%%%%%%%%%%%%%%%%%%%%%%%%
\section{Detailed Proofs}\label{app:sec:proof}

In this section, we give detailed proofs for our theoretical findings.

%%%%%%%%%%%%%%%%%%%%%%%%%%%%%%%%%
\subsection{Proof for Theorem~\ref{thrm:GST_1d_optimization}}\label{app:sec:thrm:GST_1d_optimization}

\begin{proof}
Let us consider a critic function $f\in \OrliczSobolevPsi(\G, \omega)$. Then by Definition~\ref{def:OrliczSobolev}, we have
\begin{equation}\label{eq:OSF_representation}
f(x) = f(z_0) + \int_{[z_0,x]} f'(y) \omega(\mathrm{d}y)\quad \mbox{for}\quad x\in \G.
\end{equation}
For convenience, let ${\bf{1}}_{[z_0,x]}(y)$ denote the indicator function of the shortest path $[z_0,x]$, i.e.,
\begin{equation}\label{eq:Indicator}
{\bf{1}}_{[z_0,x]}(y) = 
  \begin{cases} 
   1 & \text{if } y\in [z_0,x] \\
   0 & \text{otherwise}.
  \end{cases}
\end{equation}
Using \eqref{eq:OSF_representation}--\eqref{eq:Indicator} and due to $\mu(\G) = 1$, we can rewrite $\int_\G f(x) \mu(\mathrm{d}x)$ as 
\begin{align*}
\int_\G f(x) \mu(\mathrm{d}x) &= \int_\G f(z_0) \mu(\mathrm{d}x) + \int_\G  \int_{[z_0,x]} f'(y) \omega(\mathrm{d}y) \mu(\mathrm{d}x)\\
&= f(z_0) + \int_\G  \int_{\G}  {\bf{1}}_{[z_0,x]}(y) \, f'(y) \omega(\mathrm{d}y) \mu(\mathrm{d}x).
\end{align*}
It then follows by
applying Fubini's theorem to interchange the order of integration in the above last integral that 
\begin{align*}
\int_\G f(x) \mu(\mathrm{d}x) &= f(z_0) + \int_\G  \int_{\G}  {\bf{1}}_{[z_0,x]}(y) \, f'(y)  \mu(\mathrm{d}x) \omega(\mathrm{d}y)\\
&= f(z_0) + \int_\G  \Big[\int_{\G}  {\bf{1}}_{[z_0,x]}(y) \,  \mu(\mathrm{d}x)\Big] f'(y)  \omega(\mathrm{d}y).
\end{align*}
Owing to  the definition of $\Lambda(y)$ in \eqref{sub-graph},  the above expression is equivalent to 
\begin{align*}
\int_\G f(x) \mu(\mathrm{d}x) = f(z_0) + \int_{\G} f'(y)  \mu(\Lambda(y)) \, \omega(\mathrm{d}y).
\end{align*}

By exactly the same arguments, we also have 
\begin{align*}
\int_\G f(x) \nu(\mathrm{d}x) 
 = f(z_0) + \int_{\G} f'(y)  \nu(\Lambda(y)) \, \omega(\mathrm{d}y).
\end{align*}
As a consequence, the generalized Sobolev transport in Definition~\ref{def:GST_distance} can be rewritten as
\begin{align}\label{eq:reformulation}
\calGS_{\Phi}(\mu,\nu )  = 
\sup \limits_{f \in \mathbb{F}}~\left| \int_{\G} f'(x) \big[ \mu(\Lambda(x)) -  \nu(\Lambda(x))\big] \, \omega(\mathrm{d}x) \right|,
\end{align}
where $\mathbb{F} \coloneqq  \Big\{ f\in \OrliczSobolevPsi(\G, \omega):  \, \|f'\|_{L_{\Psi}}\leq 1 \Big\}$ with  $\| \cdot \|_{L_\Psi}$  being the Luxemburg norm defined in  \eqref{eq:Luxemburg_norm}. 

Next, on one hand it is clear that $\{ f': \, f \in \mathbb{F}\} \subset \{g\in L_{\Psi}(\G, \omega): \, \|g\|_{L_\Psi}\leq 1  \}$.
On the other hand, for any $g\in L_{\Psi}(\G, \omega)$ we have $g=f'$ with $f(x) \coloneqq  \int_{[z_0,x]} g(y) \omega(\mathrm{d}y)\in \OrliczSobolevPsi(\G, \omega)$. Therefore, we conclude that $\{ f': \, f\in \mathbb{F}\} = \{g\in L_{\Psi}(\G, \omega): \, \|g\|_{L_\Psi}\leq 1  \}$. As a consequence, we can rewrite identity \eqref{eq:reformulation} as 
\begin{align}\label{Holder_gst}
\calGS_{\Phi}(\mu,\nu )  =  
\sup \limits_{g\in L_{\Psi}(\G, \omega): \, \|g\|_{L_\Psi}\leq 1} \left| \int_{\G} g(x) h(x) \, \omega(\mathrm{d}x) \right|,
\end{align}
where $h(x) :=  \mu(\Lambda(x)) -  \nu(\Lambda(x))$ for all $x \in \G$.

%\tam{Definition the distance with $| \cdot |$, then apply the results of Amemiya norm and Orlicz norm to reformulate into 1-dimensional optimization problem.}

Recall from Definition~\ref{def:GST_distance} that $\Psi : \R_+ \to \R_+$ is the complementary function of $\Phi$.
%, defined as follows
%\begin{equation*}
%\Phi(t) = \sup \left[at - \Psi(a) \, : \, a \ge 0  \right],
%\end{equation*}
Therefore, we obtain from \eqref{Holder_gst}
and \citep[Proposition 10, pp.81]{rao1991theory} that 
\begin{align}\label{eq:same_as_Orlicz}
\calGS_{\Phi}(\mu,\nu )  =  \|h\|_{\Phi}
\end{align}
with $\|h\|_{\Phi}$ being the Orlicz norm 
define by (see \citep[Definition~2, pp.58]{rao1991theory})
\begin{equation}\label{eq:OrliczNorm}
\|h\|_{\Phi} := \sup{\Big\{ \int_{\G} | h(x) g(x)| \omega(\text{d}x):\, \int_{\G} \Psi(|g(x)|) \omega(\text{d}x) \leq 1\Big\} }. 
\end{equation}

%and observe that from \eqref{Holder_gst}, $\calGS_{\Phi}(\mu,\nu)$ is the Orlicz norm with respect to the $\Psi$ function for duality in Orlicz space~\citep[Proposition 10, pp.81]{rao1991theory},~\citep[\S8.17]{adams2003sobolev}, denoted as $\norm{h}_{\Phi}^{O}$.

%Next, applying \citep[Theorem 1]{hudzik2000amemiya}, we can rewrite $\calGS_{\Phi}(\mu,\nu)$ as follows:

By applying~\citep[Theorem 13, pp.69]{rao1991theory}, we have 
\begin{align*}
\|h\|_{\Phi}  =  
\inf_{k > 0} \frac{1}{k}\left( 1 + \int_{\G} \Phi(k \left| h(x) \right|) \omega(\text{d}x) \right).
\end{align*}
This together with \eqref{eq:same_as_Orlicz} yields
\begin{align*}
\calGS_{\Phi}(\mu,\nu )  =  
\inf_{k > 0} \frac{1}{k}\left( 1 + \int_{\G} \Phi(k \left| h(x) \right|) \omega(\text{d}x) \right).
\end{align*}
This completes the proof of the theorem.
\end{proof}

%%%%%%%%%%%%%%%%%%%%%%%%%%%%
%%%%%%%%%%%%%%%%%%%%%%%%%%%%
\subsection{Proof for Corollary~\ref{cor:GST_1d_optimization_discrete}}\label{app:subsec:cor:GST_1d_optimization_discrete}
\begin{proof}
By Theorem~\ref{thrm:GST_1d_optimization}, we have
\begin{align}\label{eq:fromTheorem3.3}
\calGS_{\Phi}(\mu,\nu )  =  
\inf_{k > 0} \frac{1}{k}\left( 1 + \int_{\G} \Phi(k \left| h(x) \right|) \omega(\text{d}x) \right)
\end{align}
with $h(x) :=  \mu(\Lambda(x)) -  \nu(\Lambda(x))$.
We next compute the integral in \eqref{eq:fromTheorem3.3}. 

Give two data points $u, v$ in $\R^n$, let $\langle u,  v\rangle$ be the line segment in $\R^n$ connecting the two points $u, v$, and denote $( u, v)$ as the same line segment but without its two end-points.

Using the assumption $\omega(\{x\})= 0$ for every $x\in\G$, we have
\[
\int_{\G} \Phi(k \left| h(x) \right|) \omega(\text{d}x) = \sum_{e=\langle u,v\rangle\in E}   \int_{(u,v)} \Phi(k \left| h(x) \right|) \omega(\text{d}x).
\]
As the input measures $\mu$ and $\nu$ are assumed to be supported on nodes $V$ of graph $\G$, we also have
$h(x) =  \mu(\Lambda(x)) -  \nu(\Lambda(x))= \mu(\Lambda(x)\setminus (u,v)) -  \nu(\Lambda(x)\setminus (u,v))$ for every edge $e=\langle u,v\rangle\in E$. Therefore,  the above identity 
can be further rewritten as
\begin{equation}\label{eq:remove_interior}
\int_{\G} \Phi(k \left| h(x) \right|) \omega(\text{d}x) = \sum_{e=\langle u,v\rangle\in E}   \int_{(u,v)} \Phi(k \left| \mu(\Lambda(x)\setminus (u,v)) -  \nu(\Lambda(x)\setminus (u,v)) \right|) \omega(\text{d}x).
\end{equation}

Consider edge $e$ between two nodes $u, v \in V$ of graph $\G$, i.e., $e=\langle u,v\rangle$. For $x\in (u,v)$, we have $y\in \G\setminus (u,v)$ belongs to $\Lambda(x)$ if and only if $y\in \gamma_e$ (see Equation~\eqref{sub-graph} for the definitions of $\Lambda(x)$ and $\gamma_e$). It follows that $\Lambda(x)\setminus (u,v) =\gamma_e$, and thus we deduce from \eqref{eq:remove_interior} that
\begin{align*}
\int_{\G} \Phi(k \left| h(x) \right|) \omega(\text{d}x) 
&= \sum_{e=\langle u,v\rangle\in E}   \int_{(u,v)} \Phi(k \left| \mu(\gamma_e) -  \nu(\gamma_e) \right|) \omega(\text{d}x) \\
&= \sum_{e=\langle u,v\rangle\in E} \Phi(k \left| \mu(\gamma_e) -  \nu(\gamma_e) \right|) \int_{(u,v)} \omega(\text{d}x)\\
&= \sum_{e \in E} w_e \, \Phi(k \left| \mu(\gamma_e) -  \nu(\gamma_e) \right|). \label{app:eq:sum_edge_on_graph}
\end{align*}
This together with \eqref{eq:fromTheorem3.3} yields
\begin{equation*}
\calGS_{\Phi}(\mu,\nu )  =  
\inf_{k > 0} \frac{1}{k}\left( 1 + \sum_{e \in E} w_e \, \Phi(k \left| \mu(\gamma_e) -  \nu(\gamma_e) \right|)    \right).
\end{equation*}
%where $\bar{h}(e) = \mu(\gamma_e) - \nu(\gamma_e)$ for any edge $e \in E$.
%in $\G$.
Thus,  the proof is  complete. 
\end{proof}

\subsection{Proof for Theorem~\ref{thrm:metrization}}
\label{app:subsec:thrm:metrization}

\begin{proof}
We will prove that the generalized Sobolev transport $\calGS_{\Phi}(\mu,\nu)$ satisfies: (i) nonnegativity, (ii) indiscernibility, (iii) symmetry, and (iv) triangle inequality.

%(ii) indiscernibility/identity

\textbf{(i) Nonnegativity.} By choosing $f=0$ in Definition~\ref{def:GST_distance},  we see that $\calGS_{\Phi}(\mu,\nu) \ge 0$ for every $(\mu, \nu)$ in $\calP(\G) \times \calP(\G)$. Therefore, the generalized Sobolev transport is nonnegative.

%\textbf{(ii) Identity.}

\textbf{(ii) Indiscernibility.} Assume that $\calGS_{\Phi}(\mu,\nu) =0$, then we have 
\begin{align}\label{int_identity}
\int_\G f(x) \mu(\mathrm{d}x) - \int_\G f(x) \nu(\mathrm{d}x) = 0,
\end{align}
for all $f \in {\OrliczSobolevPsi}(\G, \omega)$ satisfying the constraint $\|f'\|_{L_{\Psi}}\leq 1$.

 %Indeed, if there exists $\tilde f \in {\OrliczSobolevPsi}(\G, \omega)$ and such that $\|\tilde f'\|_{L_{\Psi}(\G, \omega)}\leq 1$, and 
 %\[
 %\int_\G \tilde f(x) \mu(\mathrm{d}x) - \int_\G \tilde f(x) \nu(\mathrm{d}x) < 0.
% \]
% Then, by taking $f=-\tilde f $ in Definition~\ref{def:distance}, we see that $\calS_p(\mu,\nu )> 0$ which contradicts the assumption $\calS_p(\mu,\nu ) =0$. Thus \eqref{int_identity} holds. 

Now let $g \in {\OrliczSobolevPsi}(\G, \omega)$ be any nonconstant function. Then $c := \|g'\|_{L_{\Psi}} > 0$. Then by taking $f : = \frac{g}{c}$, we have 
$f\in {\OrliczSobolevPsi}(\G, \omega)$ with  $ \|f'\|_{L_{\Psi}} =\|\frac{g'}{c}\|_{L_{\Psi}}= \frac1c \|g'\|_{L_{\Psi}} = 1$. Hence, it follows from \eqref{int_identity} that
\begin{align*}
\int_\G \frac{g(x)}{c} \mu(\mathrm{d}x) - \int_\G \frac{g(x)}{c} \nu(\mathrm{d}x) = 0,
\end{align*}
which implies that
\begin{align}\label{eq: Indiscernibility}
\int_\G g(x) \mu(\mathrm{d}x) =\int_\G g(x)\nu(\mathrm{d}x).
\end{align}
Thus, we have shown that \eqref{eq: Indiscernibility} holds true for every nonconstant function $g \in {\OrliczSobolevPsi}(\G, \omega)$. But \eqref{eq: Indiscernibility} is also obviously true for any constant function $g$. So, we in fact obtain 
\[
\int_\G g(x) \mu(\mathrm{d}x) = \int_\G g(x) \nu(\mathrm{d}x) 
\qquad \mbox{for every}\quad g\in {\OrliczSobolevPsi}(\G, \omega),
\]
which gives $\mu =\nu$ as desired. 

\textbf{(iii) Symmetry.} This property is obvious from  Definition~\ref{def:GST_distance} as the value $\calGS_{\Phi}(\mu,\nu)$ is unchanged when the role of $\mu$ and $\nu$ is interchanged. That is,   $\calGS_{\Phi}(\mu,\nu) = \calGS_{\Phi}(\nu,\mu)$.

%To prove the symmetry of $\calS_p(\mu,\nu )$, observe that if $f\in W^{1,p'}(\G, \lambda)$ with $\|f'\|_{L^{p'}(\G, \lambda)}\leq 1$, then we also have $-f \in W^{1,p'}(\G, \lambda)$ with $\|-f'\|_{L^{p'}(\G, \lambda)} = \|f'\|_{L^{p'}(\G, \lambda)} \leq 1$. As a consequence, $\calS_p(\mu,\nu ) = \calS_p(\nu,\mu)$.
%is symmetric. 

\textbf{(iv) Triangle inequality.} Let $\mu,\nu,\sigma$ be probability measures in $\calP(\G)$.  Then for any function $f \in {\OrliczSobolevPsi}(\G, \omega)$ satisfying $\|f'\|_{L_{\Psi}}\leq 1$, we have
\begin{align*}
\left| \int_\G f(x) \mu(\mathrm{d}x) - \int_\G f(x) \nu(\mathrm{d}x) \right| &= \left| \left[ \int_\G f(x) \mu(\mathrm{d}x) - \int_\G f(x) \sigma(\mathrm{d}x) \right]  + \left[\int_\G f(x) \sigma(\mathrm{d}x) - \int_\G f(x) \nu(\mathrm{d}x)\right] \right|\\
&\leq  \left|  \int_\G f(x) \mu(\mathrm{d}x) - \int_\G f(x) \sigma(\mathrm{d}x) \right|  + \left|\int_\G f(x) \sigma(\mathrm{d}x) - \int_\G f(x) \nu(\mathrm{d}x) \right|\\
&\leq \calGS_{\Phi}(\mu,\sigma) + \calGS_{\Phi}(\sigma, \nu).
\end{align*}
By taking the supremum over $f$, this implies that $\calGS_{\Phi}(\mu, \nu) \leq \calGS_{\Phi}(\mu,\sigma) + \calGS_{\Phi}(\sigma, \nu)$.

Due to the above properties, we conclude that the generalized Sobolev transport $\calGS_{\Phi}(\mu,\nu)$ is a metric on the space $\calP(\G)$ of probability measures on graph $\G$.

\end{proof}

%%%%%%%%%%%%%%%%%%%%%%%%%
\subsection{Proof for Proposition~\ref{prop:strong_metric}}\label{app:subsec:prop:strong_metric}

\begin{proof}
Let $\Psi_1$ and $\Psi_2$ be respectively the complementary functions of $\Phi_1$ and $\Phi_2$ according to definition  \eqref{eq:complementary_func}.

Let  $t\in \R_+$ be arbitrary. Since $\Phi_1 \leq \Phi_2$ we have:
\begin{eqnarray*}
   % \Phi_1(a) &&\le \quad \Phi_2(a), \\ 
   % \Rightarrow 
   at - \Phi_1(a) &&\ge \quad at - \Phi_2(a) \, \, \text{ for every } {a \in \R_+}, \\
    \Rightarrow \sup_{a \ge 0} \left( at - \Phi_1(a) \right) &&\ge \quad \sup_{a \ge 0} \left( at - \Phi_2(a) \right).
\end{eqnarray*}
%Let $\Psi_1, \Psi_2$ be the complementary functions of $\Phi_1, \Phi_2$ respectively, then
This implies that
\[
\Psi_1(t) \ge \Psi_2(t), \text{ for all } t \in \R_+.
\]
It follows that $L_{\Psi_1}(\G, \omega) \subset L_{\Psi_2}(\G, \omega)$ and ${\OrliczSobolevPsiOne}(\G, \omega) \subset {\OrliczSobolevPsiTwo}(\G, \omega)$.
Moreover, for any fixed Orlicz function $f'$ and any number $t>0$, 
%and for any nonnegative Borel measure $\omega$, 
we have
\[
\int_{\G} \Psi_1\left(\frac{|f'(x)|}{t}\right)\omega(\text{d}x) \ge \int_{\G} \Psi_2\left(\frac{|f'(x)|}{t}\right)\omega(\text{d}x).
\]
Consequently, we obtain
\[
\left\{ t >0  \mid \int_{\G} \Psi_1\left(\frac{|f'(x)|}{t}\right)\omega(\text{d}x) \le 1 \right\} \subset \left\{ t>0 \mid \int_{\G} \Psi_2\left(\frac{|f'(x)|}{t}\right)\omega(\text{d}x) \le 1 \right\}.
\]
Since the infimum of a set is smaller than or equal to the infimum of its subset, we deduce that
\[
\|f'\|_{L_{\Psi_1}} \ge \|f'\|_{L_{\Psi_2}}. 
\]
It follows from this and ${\OrliczSobolevPsiOne}(\G, \omega) \subset {\OrliczSobolevPsiTwo}(\G, \omega)$ that 
\[
\left\{f \mid f \in {\OrliczSobolevPsiOne}(\G, \omega),  \, \|f'\|_{L_{\Psi_1}(\G, \omega)}\leq 1 \right\} \subset \left\{f \mid f \in {\OrliczSobolevPsiTwo}(\G, \omega),  \, \|f'\|_{L_{\Psi_2}(\G, \omega)}\leq 1 \right\}.
\]
Since the supremum of a set is larger than or equal to the supremum of its subset, we conclude that
\[
\calGS_{\Phi_1}(\mu,\nu ) \le \calGS_{\Phi_2}(\mu,\nu )
\]
for any input measures $\mu, \nu$ in $\calP(\G)$. Therefore, the proof is completed.

We further note that one can directly leverage the result in Theorem~\ref{thrm:GST_1d_optimization}, where generalized Sobolev transport can be computed by solving univariate optimization problem, to simplify the proof.
\end{proof}

%%%%%%%%%%%%%%%%%%%%%%%%%%
\subsection{Proof for Proposition~\ref{prop:Connection_OS_Sobolev}}\label{app:subsec:prop:Connection_OS_Sobolev}

\begin{proof}

For the $N$-function $\Phi(t) = t^p$ with $1 < p < \infty$, we have 
\[
L_{\Phi}(\G, \omega) = L^p(\G, \omega),
\]
where $L^p(\G, \omega)$ is the standard $L^p$ functional space which is reviewed in \S\ref{appsec:Review_SobolevTransport} (appendix). 

Thus, following Definition~\ref{def:OrliczSobolev} and the definition of graph-based Sobolev transport~\citep[Definition 3.1]{le2022st} ( a review is given in Definition~\ref{def:Sobolev} (appendix)), we have 
\[
\OrliczSobolevPhi(\G, \omega) = W^{1, p}(\G, \omega).
\]
Thus, the proof is complete.

\end{proof}

%%%%%%%%%%%%%%%%%%%%%%%%
\subsection{Proof for Proposition~\ref{prop:Connection_GST_ST}}\label{app:subsec:prop:Connection_GST_ST}

%$\Phi(t) = \frac{1}{p}\left(\frac{1}{q}\right)^{\frac{p}{q}} t^p$ such that $1 < p < \infty$ and $q$ is a conjugate of $p$, i.e., $\frac{1}{p} + \frac{1}{q} = 1$
\begin{proof}
Let $h(x) :=  \mu(\Lambda(x)) -  \nu(\Lambda(x))$ for $x\in\G$.
Then since $\Phi(t) = \frac{(p-1)^{p-1}}{p^p} t^p$ with $1 < p < \infty$, we obtain from Theorem~\ref{thrm:GST_1d_optimization} that 
\begin{align}\label{eq:limF}
\calGS_{\Phi}(\mu,\nu )  &=  
\inf_{k > 0} \frac{1}{k}\left( 1 + \int_{\G} \frac{(p-1)^{p-1}}{p^p} k^p \left| h(x) \right|^p \omega(\text{d}x) \right) \nonumber\\
&= \inf_{k > 0} F(k),
\end{align}
where $F(k) := \frac{1}{k} + \frac{(p-1)^{p-1}}{p^p} k^{p-1} \int_{\G} \left| h(x) \right|^p \omega(\text{d}x)$ for $k>0$.

Let us consider the following two possibilities:

{\bf Case 1:} $\int_{\G} \left| h(x) \right|^p \omega(\text{d}x) = 0$. In that case, we have from \eqref{eq:limF} that
\begin{align*}
\calGS_{\Phi}(\mu,\nu ) = \inf_{k > 0} \frac1k = 0 =    \left(\int_{\G} \left| h(x) \right|^p \omega(\text{d}x)\right)^{\frac{1}{p}}= \calS_p(\mu,\nu ).
\end{align*}

{\bf Case 2:} $\int_{\G} \left| h(x) \right|^p \omega(\text{d}x) \neq 0$.
Then  $\lim_{k\to 0^+} F(k) = \lim_{k\to +\infty} F(k) = +\infty$. Therefore, it follows from  \eqref{eq:limF} that 
\begin{align}\label{eq:F(k_0)}
\calGS_{\Phi}(\mu,\nu ) = F(k_0)
\end{align}
for some finite number $k_0 \in (0, +\infty) $ satisfying $F'(k_0) =0$. As $F'(k)= -\frac{1}{k^2} + \Big(\frac{p-1}{p}\Big)^p k^{p-2} \int_{\G} \left| h(x) \right|^p$, we can solve the equation $F'(k_0) =0$ for $k_0$ to obtain 
\begin{align*}\label{eq:k_opt}
k_0 = \frac{1}{ \frac{p-1}{p}  \left(\int_{\G} \left| h(x) \right|^p \omega(\text{d}x)\right)^{\frac{1}{p}} }. 
\end{align*}
Plugging this value of $k_0$ into Equation~\eqref{eq:F(k_0)} and using the formula for $F(k)$, we get
\begin{align*}
\calGS_{\Phi}(\mu,\nu ) 
&= \frac{1}{k_0}\left( 1  + \frac{(p-1)^{p-1}}{p^p} k_0^p \int_{\G} \left| h(x) \right|^p\right)\\
&= \frac{p-1}{p}  \left(\int_{\G} \left| h(x) \right|^p \omega(\text{d}x)\right)^{\frac{1}{p}} \left( 1 + \frac{(p-1)^{p-1}}{p^p} \frac{1}{ \frac{(p-1)^p}{p^p}  \left(\int_{\G} \left| h(x) \right|^p \omega(\text{d}x)\right) } \int_{\G} \left| h(x) \right|^p \omega(\text{d}x) \right) \\
&= \left(\int_{\G} \left| h(x) \right|^p \omega(\text{d}x)\right)^{\frac{1}{p}} \\
&=  \calS_p(\mu,\nu ).
\end{align*}

Thus, we have shown that $\calGS_{\Phi}(\mu,\nu )  =  \calS_p(\mu,\nu )$ in both cases, and hence the proof is complete.
\end{proof}

\begin{remark}\label{rm:complementaryLp}
For the $N$-function $\Phi(t) = t^q$ where $1 < q < \infty$, its complementary function is $\Psi(t) = \frac{(p-1)^{p-1}}{p^p} t^p$ where $p$ is the conjugate of $q$, i.e., $\frac{1}{q} + \frac{1}{p} = 1$.
\end{remark}

%%%%%%%%%%%%%%%%%%%%%%%%%%%%%%%%%%%%%
\subsection{Proof for $\lim_{p \to 1^+} \frac{(p-1)^{p-1}}{p^p} t^p = t$ in Remark~\ref{rm:OW-TW}}\label{app:subsec:rm:OW-TW}

We will prove that $\lim_{p \to 1^+} \frac{(p-1)^{p-1}}{p^p} t^p = t$ which is used in Remark~\ref{rm:OW-TW}. For this and since $\lim_{p \to 1^+} \frac{t^p}{p^p} = t$, it is enough to show that 
\[
\lim_{p \to 1^+} (p-1)^{p-1} = 1.
\]
By taking logarithm, this in turn is equivalent to proving that
\begin{equation}\label{eq:ln}
\lim_{p \to 1^+} \ln {(p-1)^{p-1}} = 0.
\end{equation}
\begin{proof}[Proof of \eqref{eq:ln}]
%\begin{align*}
%\lim_{p \to 1} \frac{(p-1)^{p-1}}{p^p} t^p &= \lim_{p \to 1} \frac{t^p}{{p^p}} \exp{\left(\ln{\left[(p-1)^{p-1}\right]}\right)} \\
%&= \lim_{p \to 1} \frac{t^p}{{p^p}} \exp{\left[(p-1)\ln{(p-1)}\right]} \\
%&= \lim_{p \to 1} \frac{t^p}{{p^p}} \exp{\left[\frac{\ln{(p-1)}}{1/(p-1)}\right]} \\
%&= \lim_{p \to 1} \frac{t^p}{{p^p}} \exp{\left[\frac{1/(p-1)}{-1/(p-1)^2}\right]} \\
%&= \lim_{p \to 1} \frac{t^p}{{p^p}} \exp{(1-p)} \\
%&= t,
%\end{align*}
\begin{align*}
  \lim_{p \to 1^+} \ln {(p-1)^{p-1}} =   \lim_{p \to 1^+}\frac{\ln (p-1)}{1/(p-1)}
  = \lim_{p \to 1^+} \frac{1/(p-1)}{-1/(p-1)^2},
\end{align*}
where we apply the L'Hopital's rule for the last  equality. It follows that $\lim_{p \to 1^+} \ln {(p-1)^{p-1}} = \lim_{p \to 1^+} - (p-1) = 0$ as desired. Hence,  the proof is complete.
\end{proof}

\subsection{Complementary function for $\Phi(t) = \exp(t^p) - 1$ with $1 < p <\infty$}\label{app:subsec:complementary_func_exp}

For the given $N$-function $\Phi(t) = \exp(t^p) - 1$ with $1 < p < \infty$, its complementary function $\Psi$ is computed as follow
\begin{equation}\label{eq:obj_complementary_func_exp}
    \Psi(t) = \sup_{a \ge 0} \left( at - \Phi(a) \right).
\end{equation}
Let us fix $t\geq 0$ and denote the objective function in Equation~\eqref{eq:obj_complementary_func_exp} as $F(a)$. Then  $F(a) = at - \exp(a^p) + 1$, and  we have
\begin{eqnarray*}
    && F'(a) = t - p a^{p-1} \exp(a^p), \\
    && F''(a) = - p (p-1) a^{p-2} \exp(a^p) - p^2 a^{2(p-1)}\exp(a^p)
    = -\Big[p (p-1) a^{p-2} + p^2 a^{2(p-1)} \Big] \exp(a^p).
\end{eqnarray*}
Since $1 < p < \infty$,  we obtain $F''(a) < 0$ for all $a > 0$. Thus, $F'(a)$ is a strictly decreasing function on $(0, +\infty)$. Additionally, we have
\begin{eqnarray*}
    && F'(0) = t\geq 0, \\
    && \lim_{a \to +\infty} F'(a) = -\infty.
\end{eqnarray*}
Therefore, $F'(a) = 0$ has a unique root,  which is denoted by $a^*$. Then, the complementary function $\Psi$  is
\[
\Psi(t) =  \sup_{a \ge 0} F(a) = F(a^*) = a^{*} t - \exp((a^{*})^p) - 1.
\]

%%%%%%%%%%%%%%%%%%%%%%%%%%%%%%%%%%%%%%%%%%%%
%%%%%%%%%%%%%%%%%%%%%%%%%%%%%%%%%%%%%%%%%%%%
\section{Further Results and Discussions}\label{app:sec:further_results_discussions}

In this section, we give further results and discussions.

\textbf{Notations.} Let $\langle x, y\rangle$ be the line segment in $\R^n$ connecting two points $x, y$, and denote $(x, y)$ as the same line segment but without its two end-points. 

%%%%%%%%%%%%%%%%%%%%%%%%%%%%%%%%%%%%%
%%%%%%%%%%%%%%%%%%%%%%%%%%%%%%%%%%%%%
\subsection{On Sobolev transport}\label{appsec:Review_SobolevTransport}

%%%%%%%%%%%%%%%%%%%%%%%%%%%%%%%%%%%%%
\textbf{$\boldsymbol{L^{p}}$ functional space.} For a nonnegative Borel measure $\omega$ on $\G$, denote $L^p( \G, \omega)$ as the space of all Borel measurable functions $f:\G\to \R$ such that $\int_\G |f(y)|^p \omega(\mathrm{d}y) <\infty$. For $p=\infty$, we instead assume that $f$ is bounded $\omega$-a.e. 

Functions $f_1, f_2 \in L^p( \G, \omega)$ are considered to be the same if $f_1(x) =f_2(x)$ for $\omega$-a.e. $x\in\G$. 

Then, $L^p( \G, \omega)$ is a normed space with the norm defined by
\[
\|f\|_{L^p(\G, \omega)} := \left(\int_\G |f(y)|^p \omega(\dd y)\right)^\frac{1}{p} \text{ for } 1\leq p < \infty, \text{ and}
\]
\[
\|f\|_{L^{\infty}(\G, \omega)} := \inf\left\{t \in \R:\, |f(x)|\leq t \mbox{ for $\omega$-a.e. } x\in\G\right\}.
\]

%%%%%%%%%%%%%%%%%%%%%%%%%%%%%%%%%%%%%
\textbf{Connection between $\boldsymbol{L^{p}}$ and $\boldsymbol{L_{\Phi}}$ functional spaces.} When the convex function $\Phi(t) = t^p$, for $1 < p < \infty$, we have 
\[
L^{p}(\G, \omega) = L_{\Phi}(\G, \omega).
\]

%%%%%%%%%%%%%%%%%%%%%%%%%%%%%%%%%%%%%
\begin{definition}[Graph-based Sobolev space~\citep{le2022st}] \label{def:Sobolev}
Let $\omega$ be a nonnegative Borel measure on $\G$, and let  $1\leq p\leq \infty$. A continuous function $f: \G \to \R$ is said to belong to the Sobolev space $W^{1,p}(\G, \omega)$ if there exists a  function $h\in L^p( \G, \omega) $ satisfying 
\begin{equation}\label{FTC}
f(x) -f(z_0) =\int_{[z_0,x]} h(y) \omega(\mathrm{d}y)  \quad \forall x\in \G.
\end{equation}
Such function  $h$ is unique in $L^p(\G, \omega) $ and is called the graph derivative of $f$ w.r.t.~the measure $\omega$. The graph derivative of $f \in W^{1,p}(\G, \omega)$ is denoted $f' \in L^p( \G, \omega)$.
\end{definition}

%%%%%%%%%%%%%%%%%%%%%%%%%%%%%%%%%%%%%
\begin{proposition}[Closed-form expression of ST~\citep{le2022st}]\label{app:prop:closed-form_ST}
Let $\omega$ be any nonnegative Borel measure on $\G$, and let  $1\leq p\leq \infty$. Then, we have 
\[
\calS_{p}(\mu,\nu )
= \left( \int_{\G} | \mu(\Lambda(x)) -  \nu(\Lambda(x))|^p \, \omega(\mathrm{d}x) \right)^{\frac{1}{p}},
\]
where $\Lambda(x)$ is the subset of $\G$  defined by Equation~\eqref{sub-graph}. 
\end{proposition}

%\textbf{$L^p$ space.} For a nonnegative Borel measure $\omega$ on $\G$, the space $L^p(\G, \omega)$ is defined as the space of all Lebesgue measurable functions $f: \G \to \R$ such that
%\[
%\norm{f}_{p} = \left[ \int_{\G} \left| f(x) \right|^p  \omega(dx)\right]^{\frac{1}{p}}
%\]

%%%%%%%%%%%%%%%%%%%%%%%%%%%%%%%%%%%%%
\begin{definition}[Length measure~\citep{le2022st}] \label{def:measure} 
Let $ \omega^*$ be the unique Borel measure on $\G$ such that the restriction of $\omega^*$ on any edge is the length measure of that edge. That is, $\omega^*$  satisfies:
\begin{enumerate}
\item[i)] For  any edge $e$ connecting two nodes $u$ and $v$, we have 
 $\omega^*(\langle x,y\rangle) = (t-s) w_e$ 
 whenever $x = (1-s) u + s v$ and $y = (1-t)u + t v$ for $s,t \in [0,1)$ with $s \leq t$. Here, $\langle x,y\rangle$ is the line segment in $e$ connecting $x$ and $y$.
 \item[ii)] For any Borel set $F \subset \G$, we have
 \[
 \omega^*(F) = \sum_{e\in E} \omega^*(F\cap e).
 \]
\end{enumerate}
\end{definition}

%%%%%%%%%%%%%%%%%%%%%%%%%%%%%%%%%%%%%
\begin{lemma}[$\omega^*$ is the length measure on graph~\citep{le2022st}] \label{lem:length-measure}
Suppose that $\G$ has no short cuts, namely, any edge $e$ is a shortest path connecting its two end-points. Then, $\omega^*$ is a length measure in the sense that
\[
\omega^*([x,y]) = d_\G(x,y)
\]
for  any  shortest path   $[x,y]$ connecting $x, y$. Particularly, $\omega^*$ has no atom in the sense that $\omega^*(\{x\})=0$ for every $x \in \G$. 
\end{lemma}

%%%%%%%%%%%%%%%%%%%%%%%%%%%%%%%%%%%%%
%%%%%%%%%%%%%%%%%%%%%%%%%%%%%%%%%%%%%
\subsection{On Wasserstein distance}

We review the definition of the $p$-Wasserstein distance with graph metric cost for measures on graph $\G$.

\begin{definition}\label{def:pWasserstein}
Let $1\leq p <\infty$, suppose that  $\mu$ and $\nu$ are  two nonnegative Borel measures on $\G$ satisfying $\mu(\G) =\nu(\G)=1$. Then, the $p$-order Wasserstein distance between $\mu$ and $\nu$ is defined as follows:
\begin{align*}
\calW_p(\mu,\nu)^p 
&= \inf_{\gamma \in \Pi(\mu,\nu)}\int_{\G\times\G} d_\G(x,y)^p \gamma(\dd x, \dd y),
\end{align*}
where 
\[
\Pi(\mu,\nu) := \Big\{ \gamma \in \calP(\G \times \G): \, \gamma_1= \mu, \, \gamma_2= \nu \Big\},
\]
where $\gamma_1, \gamma_2$ are the first and second marginals of $\gamma$ respectively.
\end{definition}

%%%%%%%%%%%%%%%%%%%%%%%
%%%%%%%%%%%%%%%%%%%%%%%
\subsection{On generalized Sobolev transport}

\textbf{Young inequality.} Let $\Phi, \Psi$ be a pair of complementary $N$-functions, then
\[
	st \le \Psi(s) + \Phi(t).
\]

\textbf{Equivalence~\citep[\S8.17]{adams2003sobolev}~\citep[\S13.11]{musielak2006orlicz}.} The Luxemburg norm is equivalent to the Orlicz norm
\begin{equation}\label{eq:LuxemburgOrlicz}
\norm{f}_{L_\Phi} \le \norm{f}_{\Phi} \le 2 \norm{f}_{L_\Phi},
\end{equation}
where we recall that the Luxemburg norm $\norm{\cdot}_{L_\Phi}$ is defined in Equation~\eqref{eq:Luxemburg_norm}, and the Orlicz norm $\norm{\cdot}_{\Phi}$ is defined in Equation~\eqref{eq:OrliczNorm} (see \citep[Definition~2, pp.58]{rao1991theory}).

% \textbf{Other definition for Orlicz norm~\citep[\S8.17]{adams2003sobolev}~\citep{benkirane2014variational}.} Let $\Phi, \Psi$ be a pair of complementary $N$-functions, then
% \begin{equation}\label{eq:Orlicz-conjugate}
%     \norm{f}_{\Phi}^{O} = \sup \left\{ \int_{\G} f(x)g(x) \omega(dx) \mid \norm{g}_{\Psi}^{L} \le 1 \right\}.
% \end{equation}

\textbf{Generalized H\"older inequality.} Let $\Phi, \Psi$ be a pair of complementary $N$-functions, then H\"older inequality w.r.t. Luxemburg norm~\citep[\S8.11]{adams2003sobolev} is as follows:
\begin{equation}\label{eq:Holder-Luxemburg2}
\left| \int_{\G} f(x)g(x) \omega(dx) \right| \le 2 \norm{f}_{L_\Phi} \norm{g}_{L_\Psi}.
\end{equation}
Additionally, H\"older inequality w.r.t. Luxemburg norm and Orlicz norm~\citep[\S13.13]{musielak2006orlicz} is as follows:
\begin{equation}\label{eq:Holder-LuxemburgOrlicz}
\left| \int_{\G} f(x)g(x) \omega(dx) \right| \le \norm{f}_{L_\Phi} \norm{g}_{\Psi}.
\end{equation}

\subsection{Further discussions}\label{app:subsec:discussion}

For completeness, we recall important discussions on the underlying graph in~\citet{le2022st} for ST, since they are also applied and/or adapted for the GST.

%%%%%%%%%%%%%%%%%%%%%%%%%%%%%%%%%%%%%%%%%%%%%%
\textbf{OT problems for measures on a graph.} In this work, our considered problem is to compute OT distance between two input probability measures supported on the \emph{same} graph. We distinguish this line of research to the OT problem between \emph{two (different) input graphs}, considered in~\citet{petric2019got, dong2020copt, tam2022multiscale}, where its goal is to compute the distance between two input graphs by using OT approach.

\textbf{Algorithms for the univariate optimization to compute GST.} For empirical simulations, we leverage the \texttt{fmincon} in MATLAB with the \emph{trust-region-reflective} algorithm since it is easy to compute the gradient and Hessian for the objective function for the univariate optimization problem.

%%%%%%%%%%%%%%%%%%%%%%%%%%%%%%%%%%%%%%%%%%%%%%
\textbf{Path length for points in $\G$~\citep{le2023scalable}.} We can canonically measure a path length connecting any two points  $x, y \in \G$ where $x, y$ are not necessary to be nodes in $V$. Indeed, for $x, y \in \R^n$ belonging to the same edge $e= \langle u, v\rangle$ which connects two nodes $u, v \in V$, we have 
\begin{eqnarray*}
& x = (1-s) u + s v, \\
& y = (1-t)u + t v,
\end{eqnarray*}
for some numbers $t,s\in [0,1]$. Therefore, the length of the path connecting $x, y$ along the edge $e$ (i.e., the line segment $\langle x, y\rangle$) is defined by $|t-s| w_e$. 

Thus, the length for an arbitrary path in $\G$ can be similarly defined by breaking down into pieces over edges and summing over their corresponding lengths~\citep{le2022st}.

%%%%%%%%%%%%%%%%%%%%%%%%%%%%%%%%%%%%%%%%%%%%%%
\textbf{Extension to measures supported on $\G$.} Similar to ST~\citep{le2022st}, the discrete case of GST in Equation~\eqref{equ:GST_1d_optimization_discrete} can be extended for measures with finite supports on $\G$ (i.e., measures which may have supports on edges) by using the same strategy to measure a path length for points in $\G$ (discussed in the above paragraph). More precisely, we break down edges containing supports into pieces and sum over their corresponding values instead of the sum over edges as in Equation~\eqref{equ:GST_1d_optimization_discrete}.

%%%%%%%%%%%%%%%%%%%%%%%%%%%%%%%%%%%%%%%%%%%%%%
\textbf{About the assumption of uniqueness property of the shortest paths on $\G$.} As discussed in \citep{le2022st} for the ST, note that $w_e \in \R$ for any edge $e \in E$ in $\G$., it is almost surely that every node in $V$ can be regarded as unique-path root node (with a high probability, lengths of paths connecting any two nodes in graph $\G$ are different).

Additionally, for some special graph, e.g., a grid of nodes, there is \emph{no} unique-path root node for such graph. However, by perturbing each node (and/or perturbing lengths of edges in case $\G$ is a non-physical graph) with a small deviation, we can obtain a graph satisfying the unique-path root node assumption.

%%%%%%%%%%%%%%%%%%%%%%%%%%%%%%%%%%%%%%%%%%%%%%
\textbf{About the generalized Sobolev transport (GST).} Similar to the ST~\citep{le2022st}, we assume that the graph metric space (i.e., the graph structure) is given, and leave the question to learn an optimal graph metric structure from data for future work.

%%%%%%%%%%%%%%%%%%%%%%%%%%%%%%%%%%%%%%%%%%%%%%
\textbf{About graphs $\G_{\text{Log}}$ and $\G_{\text{Sqrt}}$~\citep{le2022st}.} We use a clustering method, e.g., the farthest-point clustering, to partition supports of measures into at most $M$ clusters.\footnote{$M$ is the input number of clusters used for the clustering method. Therefore, the clustering result has at most $M$ clusters, depending on input data.} Then, let $V$ be the set of centroids of these clusters, i.e., graph vertices. For edges, in graph $\G_{\text{Log}}$, we randomly choose $M\log(M)$ edges; and $M^{3/2}$ edges for graph $\G_{\text{Sqrt}}$. We further denote the set of those randomly sampled edges as $\tilde{E}$.  

For each edge $e$, its corresponding edge length (i.e., weight) $w_e$ is computed by the Euclidean distance between the two corresponding nodes of edge $e$. Let $n_c$ be the number of connected components in the graph $\tilde{\G}(V, \tilde{E})$. Then, we randomly add $(n_c - 1)$ more edges between these $n_c$ connected components to construct a connected graph $\G$ from $\tilde{\G}$. Let $E_c$ be the set of these $(n_c - 1)$ added edges and denote set $E = \tilde{E} \cup E_c$, then $\G(V, E)$ is the constructed graph. 

%\textbf{Datasets and Computational Devices.} For document dataset (i.e., \texttt{TWITTER, RECIPE, CLASSIC, AMAZON}), orbit dataset (\texttt{Orbit}) and a $10$-class subset of \texttt{MPEG7} dataset, one can contact the authors of~\citep{le2022st} to access to these datasets. For computational devices, we run all of our experiments on commodity hardware.

\textbf{Huber function and its corresponding normalized function~\citep{andoni2018subspace}.} Consider the Huber function~\citep{huber1992robust} on the domain $[0, +\infty)$, defined as, 
\begin{equation}\label{eq:Huber}
f_{H}(t) = 
  \begin{cases} 
   t^2/2 & \text{if } 0 \leq t \leq \delta \\
   \delta(t - \delta/2)       & \text{if } t > \delta,
  \end{cases}
\end{equation}
where $\delta > 0$ is a constant. Then, the normalized Huber function~\citep{andoni2018subspace} on the domain $[0, +\infty)$ is defined as
\begin{equation}\label{eq:normalizedHuber}
\hspace{-0.2em}\Phi_{H}(t) = 
  \begin{cases} 
   f_{H}(f_{H}^{-1}(1)t) & \text{if } 0 \leq t \leq 1 \\
   \Phi'_{H_{-}}(1)t - (\Phi'_{H_{-}}(1) - 1)       & \text{if } t > 1,
  \end{cases}
\end{equation}
where $\Phi'_{H_{-}}$ is the left derivative of $\Phi_{H}$.

\textbf{Hyperparamter validation.} For validation, we further randomly split \emph{the training set} into $70\%/30\%$ for validation-training and validation with $10$ repeats to choose hyper-paramters in our simulations.

\textbf{The number of pairs in training and test for kernel SVM.} Let $N_{tr}, N_{te}$ be the number of measures used for training and test respectively. For the kernel SVM training, the number of pairs which we compute the distances is $(N_{tr}-1) \times \frac{N_{tr}}{2}$. For the test phase, the number of pairs which we compute the distances is $N_{tr} \times N_{te}$. Therefore, for $1$ repeat, the number of pairs which we compute the distances for both training and test is totally $N_{tr} \times (\frac{N_{tr}-1}{2} +  N_{te})$.

%%%%%%%%%%%%%%%%%%%%%%%%%%%%%%%%%%%%%%
\subsection{Further empirical results}\label{app:subsec:further_empirical_results}

\begin{figure*}
%\begin{wrapfigure}{r}{0.22\textwidth}
%  \vspace{-6pt}
  \begin{center}
    \includegraphics[width=0.65\textwidth]{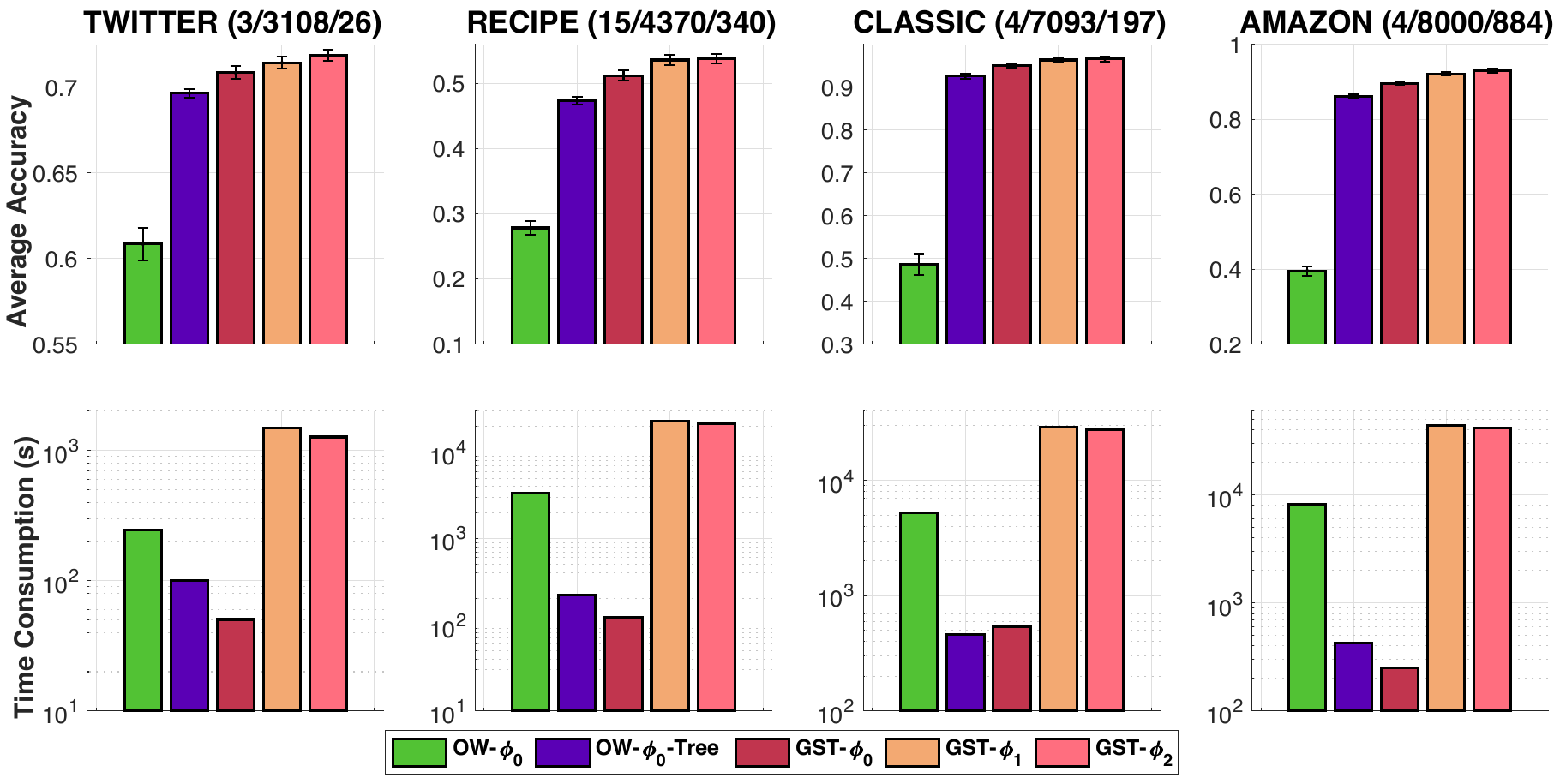}
  \end{center}
  \vspace{-6pt}
  \caption{Document classification on graph $\G_{\text{Log}}$ ($M=10^3$).}
  \label{fg:DOC_LLE_1K}
 \vspace{-6pt}
\end{figure*}

\begin{figure*}
%\begin{wrapfigure}{r}{0.22\textwidth}
%  \vspace{-6pt}
  \begin{center}
    \includegraphics[width=0.65\textwidth]{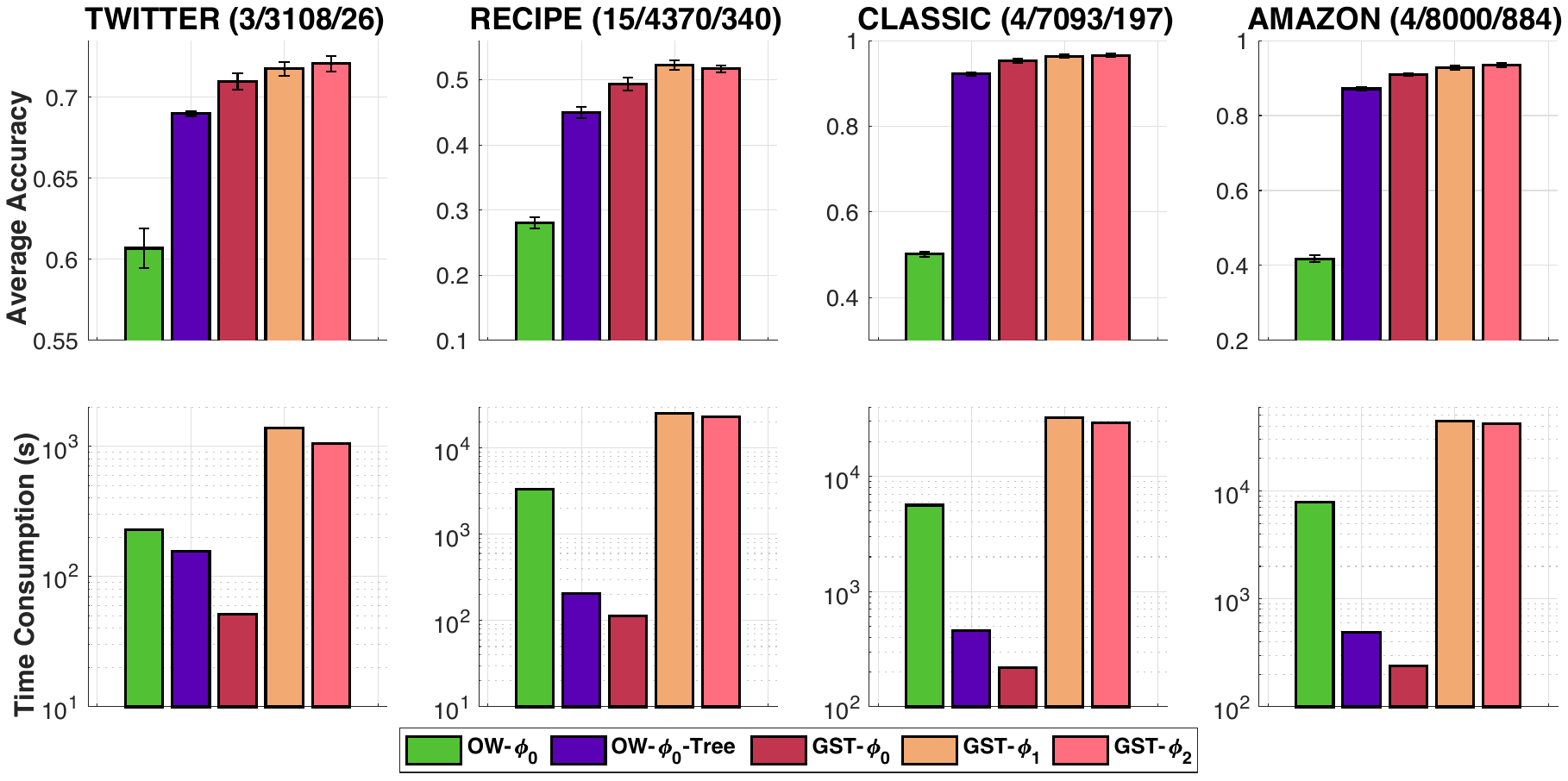}
  \end{center}
  \vspace{-6pt}
  \caption{Document classification on graph $\G_{\text{Sqrt}}$ ($M=10^3$).}
  \label{fg:DOC_SLE_1K}
 \vspace{-6pt}
\end{figure*}

\begin{figure*}
%\begin{wrapfigure}{r}{0.22\textwidth}
%  \vspace{-6pt}
  \begin{center}
    \includegraphics[width=0.65\textwidth]{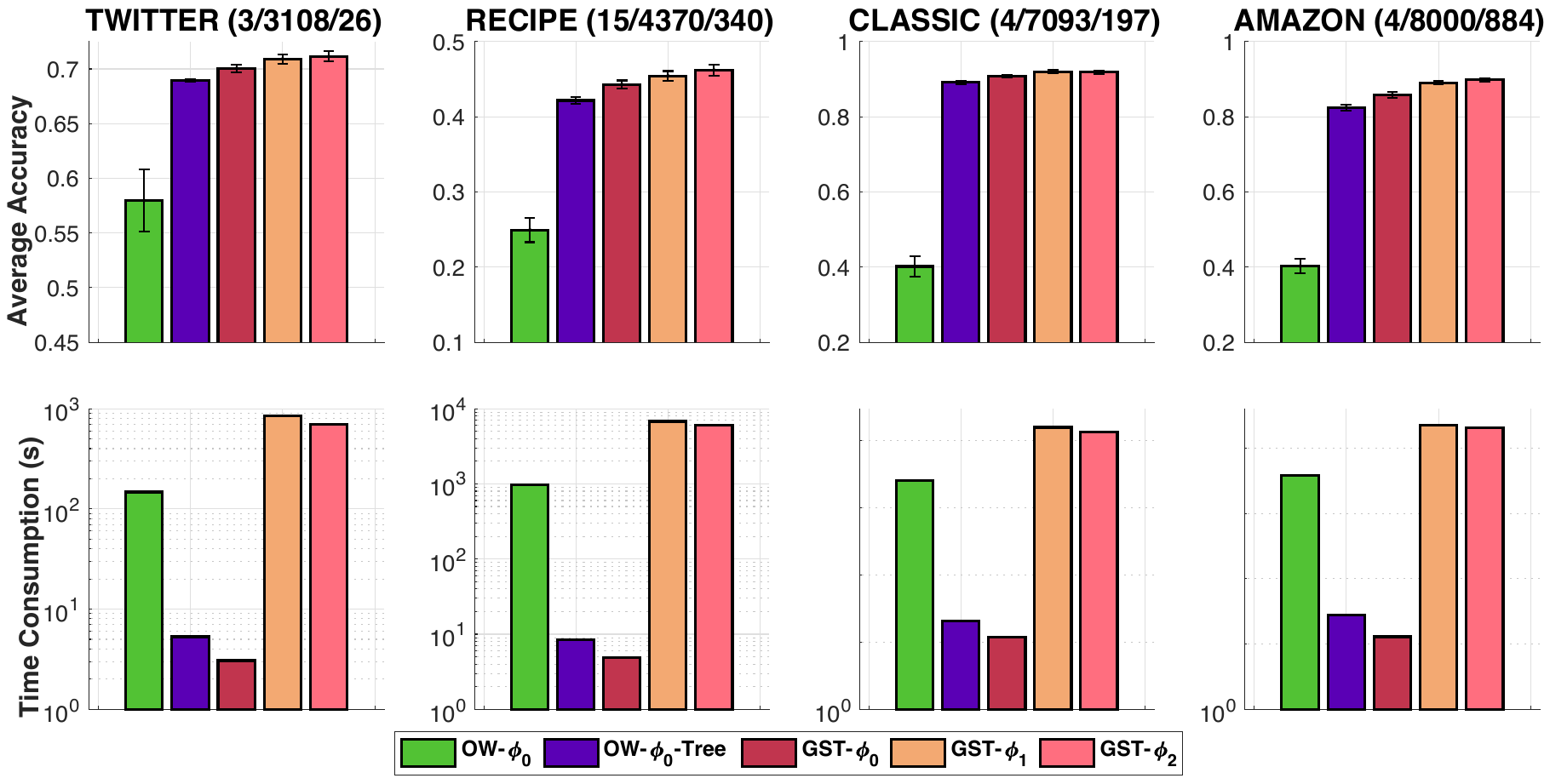}
  \end{center}
  \vspace{-6pt}
  \caption{Document classification on graph $\G_{\text{Log}}$ ($M=10^2$).}
  \label{fg:DOC_LLE_100}
 \vspace{-6pt}
\end{figure*}

\begin{figure*}
%\begin{wrapfigure}{r}{0.22\textwidth}
%  \vspace{-6pt}
  \begin{center}
    \includegraphics[width=0.65\textwidth]{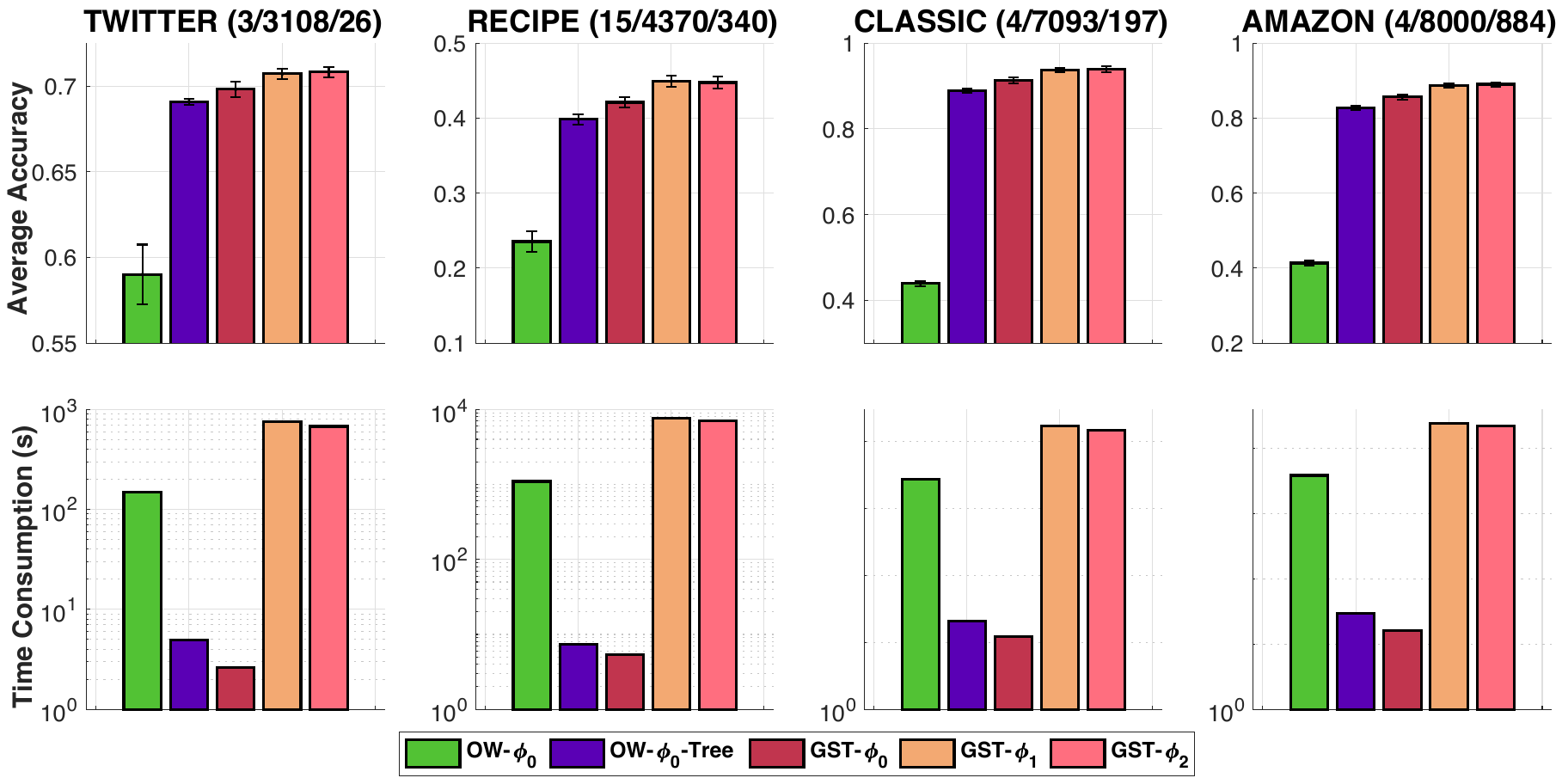}
  \end{center}
  \vspace{-6pt}
  \caption{Document classification on graph $\G_{\text{Sqrt}}$ ($M=10^2$).}
  \label{fg:DOC_SLE_100}
 \vspace{-6pt}
\end{figure*}

\textbf{Document classification.} We further illustrate document classification on graphs $\G_{\text{Log}}$ and $\G_{\text{Sqrt}}$ for:
\begin{itemize}
    \item $M = 10^3$ in Figures~\ref{fg:DOC_LLE_1K} and~\ref{fg:DOC_SLE_1K} respectively.
    \item $M = 10^2$ in Figures~\ref{fg:DOC_LLE_100} and~\ref{fg:DOC_SLE_100} respectively.
\end{itemize}

\begin{figure}
%\begin{wrapfigure}{r}{0.22\textwidth}
%  \vspace{-6pt}
  \begin{center}
    \includegraphics[width=0.35\textwidth]{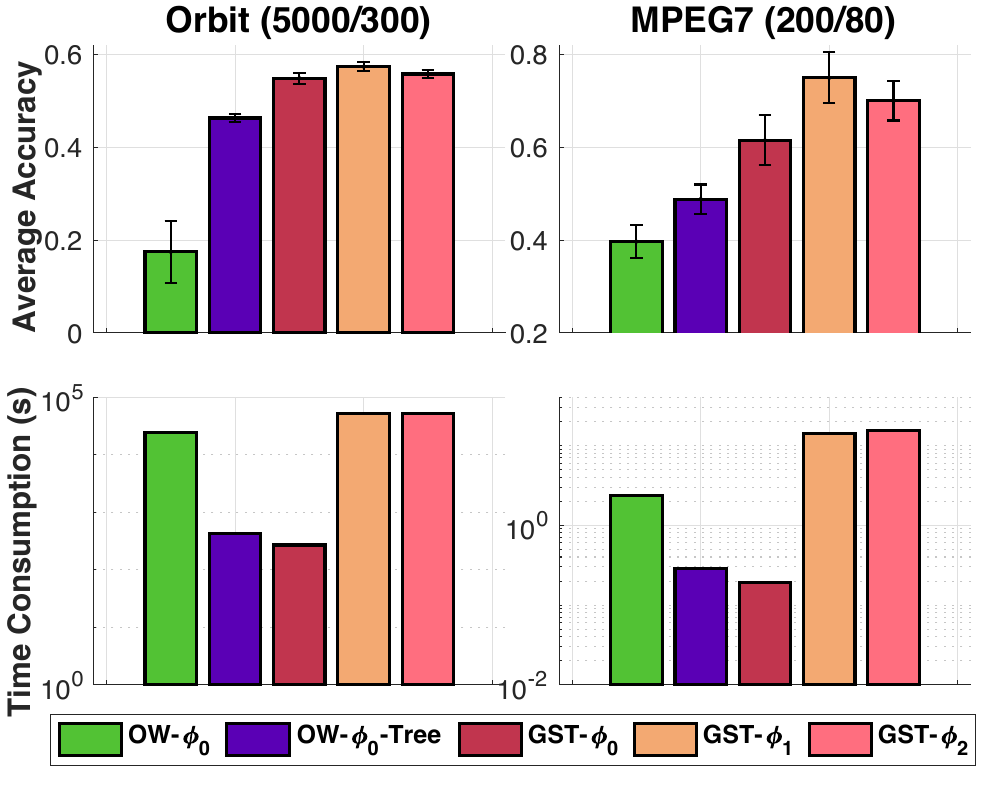}
  \end{center}
  \vspace{-6pt}
  \caption{Topological data analysis on graph $\G_{\text{Log}}$ ($M=10^3$).}
  \label{fg:TDA_LLE_1K}
 \vspace{-6pt}
\end{figure}

\begin{figure}
%\begin{wrapfigure}{r}{0.22\textwidth}
%  \vspace{-6pt}
  \begin{center}
    \includegraphics[width=0.35\textwidth]{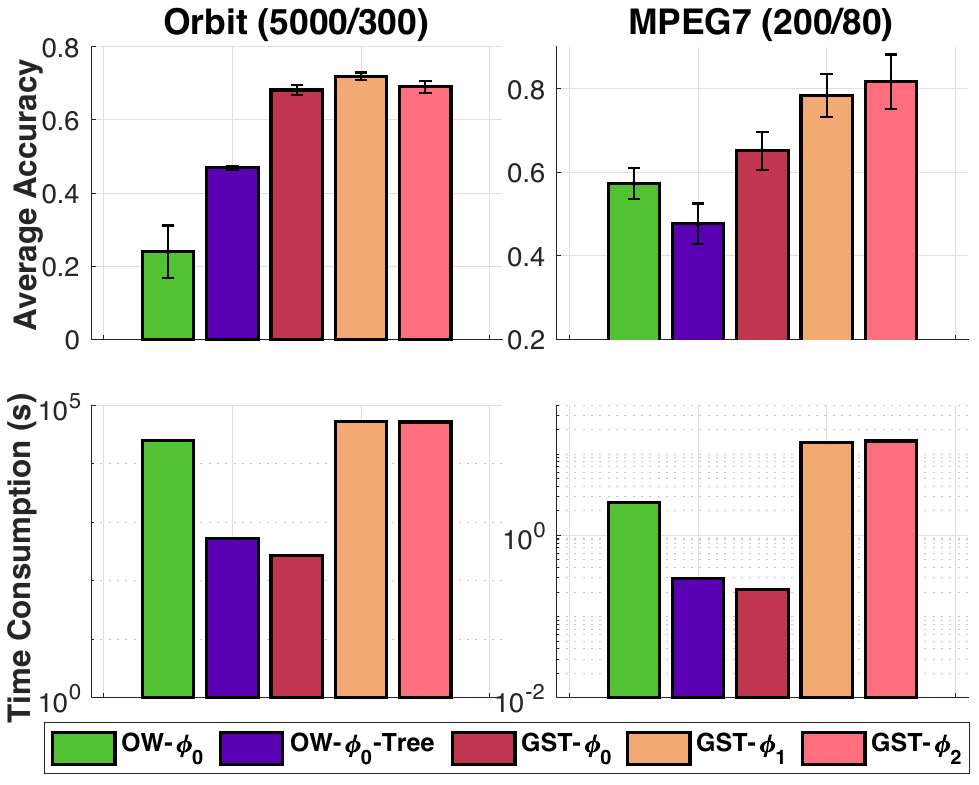}
  \end{center}
  \vspace{-6pt}
  \caption{Topological data analysis on graph $\G_{\text{Sqrt}}$ ($M=10^3$).}
  \label{fg:TDA_SLE_1K}
 \vspace{-6pt}
\end{figure}

\begin{figure}
%\begin{wrapfigure}{r}{0.22\textwidth}
%  \vspace{-6pt}
  \begin{center}
    \includegraphics[width=0.35\textwidth]{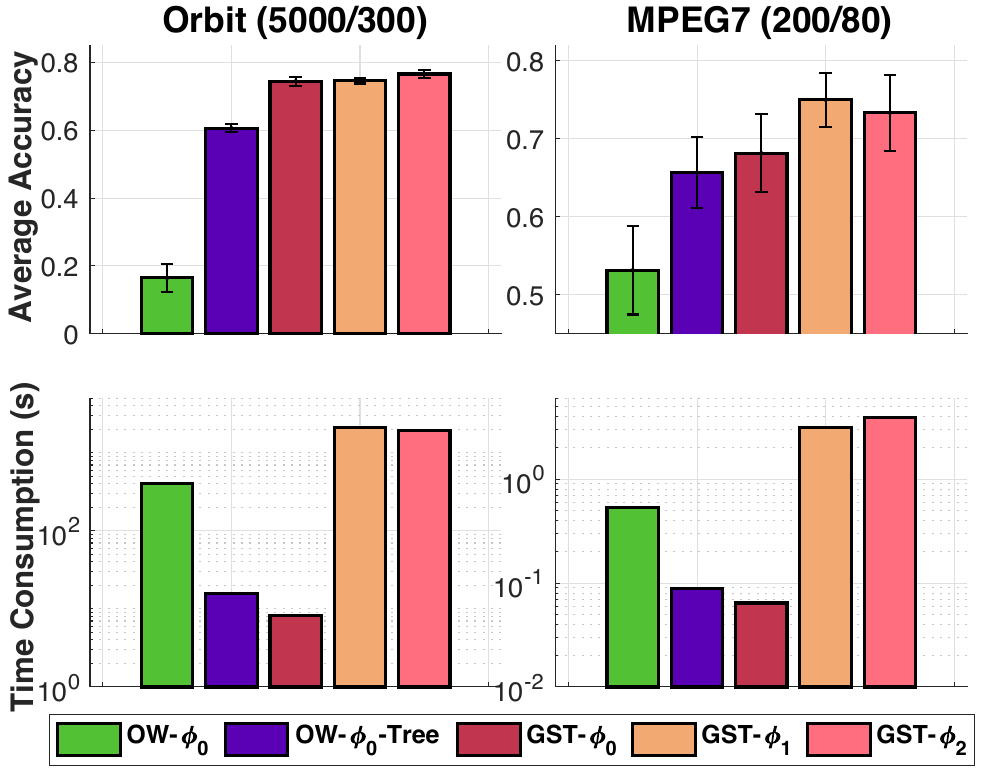}
  \end{center}
  \vspace{-6pt}
  \caption{Topological data analysis on graph $\G_{\text{Log}}$ ($M=10^2$).}
  \label{fg:TDA_LLE_100}
 \vspace{-6pt}
\end{figure}

\begin{figure}
%\begin{wrapfigure}{r}{0.22\textwidth}
%  \vspace{-6pt}
  \begin{center}
    \includegraphics[width=0.35\textwidth]{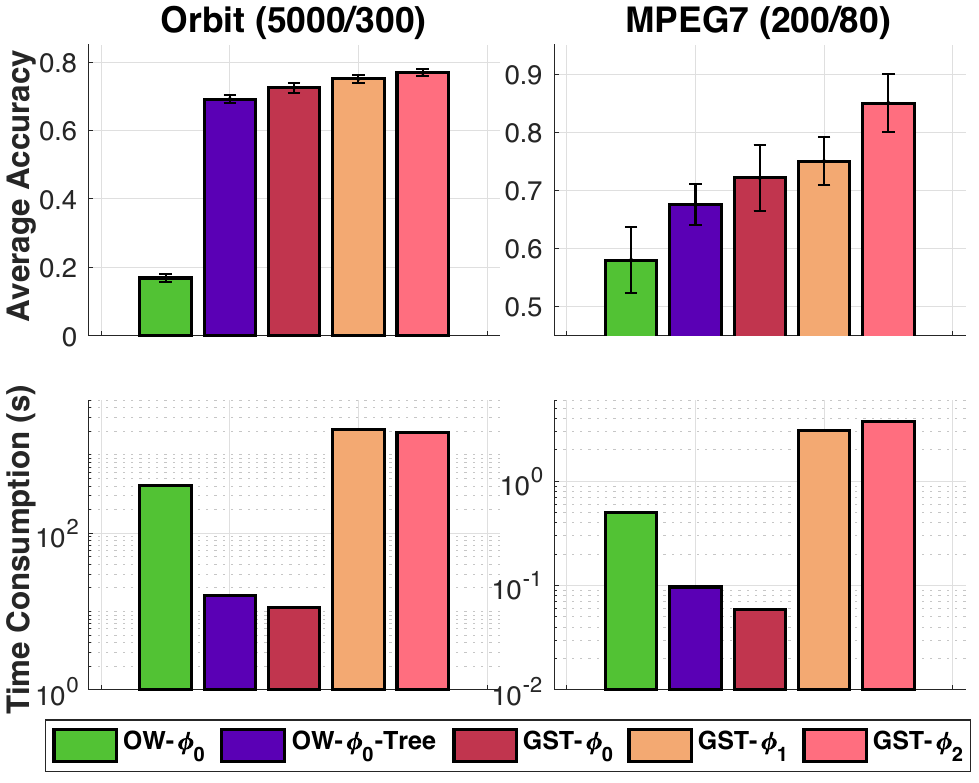}
  \end{center}
  \vspace{-6pt}
  \caption{Topological data analysis on graph $\G_{\text{Sqrt}}$ ($M=10^2$).}
  \label{fg:TDA_SLE_100}
 \vspace{-6pt}
\end{figure}

\textbf{TDA.} We also further illustrate TDA on graphs $\G_{\text{Log}}$ and $\G_{\text{Sqrt}}$ for:
\begin{itemize}
    \item $M = 10^3$ in Figures~\ref{fg:TDA_LLE_1K} and~\ref{fg:TDA_SLE_1K} respectively.
    \item $M = 10^2$ in Figures~\ref{fg:TDA_LLE_100} and~\ref{fg:TDA_SLE_100} respectively.
\end{itemize}
%%%%%%%%%%%%%%%%%%%%%%%%%%%%%%%%%%%%
% 1K
%\end{wrapfigure}

%\end{wrapfigure}

%%%%%%%%%%%%%%%%%%%%%%%%%%%%%%%%%%%%
% 100
%\end{wrapfigure}

%\end{wrapfigure}

%%%%%%%%%%%%%%%%%%%%%%%%%%%%%%%%%%
% 1K
%\end{wrapfigure}

%\end{wrapfigure}

%%%%%%%%%%%%%%%%%%%%%%%%%%%%%%%%%%
% 100
%\end{wrapfigure}

%\end{wrapfigure}

\end{document}

%% file: comments.tex
\makeatletter
\def\munderbar#1{\underline{\sbox\tw@{$#1$}\dp\tw@\z@\box\tw@}}
\makeatother

\newtheorem{theorem}{Theorem}[section]
\newtheorem{definition}[theorem]{Definition}
\newtheorem{lemma}[theorem]{Lemma}
\newtheorem{remark}[theorem]{Remark}
\newtheorem{corollary}[theorem]{Corollary}
\newtheorem{proposition}[theorem]{Proposition}

\newcommand{\be}{\begin{equation}}
\newcommand{\ee}{\end{equation}}
\newcommand{\bea}{\begin{equation*}\begin{aligned}}
\newcommand{\eea}{\end{aligned}\end{equation*}}

\newcommand{\R}{\mathbb{R}}

\newcommand{\G}{{\mathbb G}}

\newcommand{\calP}{{\mathcal P}}
\newcommand{\calS}{{\mathcal S}}

\newcommand{\calW}{{\mathcal W}}

\newcommand{\dd}{\mathrm{d}}

\newcommand{\norm}[1]{\left\lVert#1\right\rVert}